%% file: manuscript.tex
\documentclass[lettersize,journal]{IEEEtran}
\usepackage{times}

\usepackage{multicol}
\usepackage[bookmarks=true]{hyperref}

\usepackage{xparse}
\usepackage{url}
\usepackage{graphicx}
\usepackage{amsmath}
\usepackage{amssymb}

\usepackage{array}
\usepackage{balance}
\usepackage{multirow}
\usepackage{float}
\usepackage{mathrsfs}
\usepackage{listings}
\usepackage{rotating}
\usepackage{overpic}
\usepackage{subfig}

\usepackage[ruled,vlined]{algorithm2e}
\usepackage{etoolbox}%
\providetoggle{beginFlag}
\settoggle{beginFlag}{true}

\usepackage{cleveref}
\crefformat{equation}{(#2#1#3)}
\Crefname{figure}{Fig.}{Figs.}
\crefname{figure}{Fig.}{Figs.}
\crefformat{figure}{#2Fig.~#1#3}
\Crefformat{figure}{#2Fig.~#1#3}
\crefformat{table}{#2Tab.~#1#3}
\Crefformat{table}{#2Tab.~#1#3}
\crefformat{section}{#2Sec.~#1#3}
\Crefformat{section}{#2Sec.~#1#3}
\crefformat{subsection}{#2Sec.~#1#3}
\Crefformat{subsection}{#2Sec.~#1#3}
\crefformat{problem}{#2Problem~#1#3}
\Crefformat{problem}{#2Problem~#1#3}
\crefformat{algorithm}{#2Algorithm~#1#3}
\Crefformat{algorithm}{#2Algorithm~#1#3}
\crefname{appsec}{Appendix}{Appendices}

\newtheorem{assertion}{Assertion}

\usepackage{booktabs}

\usepackage{rigidnotation}
\SetConciseNotation
\input{notation-shortcuts}

\NewRigidNotation{VVec}{v}
\NewRigidNotation{UVec}{u}
\NewRigidNotation{EForce}{e}
\NewRigidNotation{RForce}{f}
\NewRigidNotation{Wrench}{w}
\NewRigidNotation{TVec}{\tau}
\NewRigidNotation{Disp}{d}
\NewRigidNotation{Twist}{\nu}
\NewRigidNotation{Acc}{a}
\def\CC{{c}}
\def\Mag{{s}}
\def\SR{{\rm{R}}}

\IEEEtriggercmd{\newpage}

\begin{document}

\title{Robustness Assessment of Assemblies\\in Frictional Contact}

\author{Philippe Nadeau and Jonathan Kelly}

\maketitle

\begin{abstract}
This work establishes a solution to the problem of assessing the capacity of multi-object assemblies to withstand external forces without becoming unstable. 
Our physically-grounded approach handles arbitrary structures made from rigid objects of any shape and mass distribution without relying on heuristics or approximations.
The result is a method that provides a foundation for autonomous robot decision-making when interacting with objects in frictional contact.
Our strategy relies on a contact interface graph representation to reason about instabilities and makes use of object shape information to decouple sub-problems and improve efficiency.
Our algorithm can be used by motion planners to produce safe assembly transportation plans, and by object placement planners to select better poses.
Compared to prior work, our approach is more generally applicable than commonly used heuristics and more efficient than dynamics simulations.
\end{abstract}

\def\abstractname{Note to Practitioners}
\begin{abstract}
This work aims at improving the autonomy of mobile manipulators in industrial object handling tasks such as planning stable object placements, and safely transporting or disassembling complex rigid object assemblies.
In practice, our algorithms can be used with assemblies of objects for which CAD models are available or have been reconstructed from sensor data.
Compared to existing methods, our approach is more accurate and much faster, executing in less than 1 second in most practical scenarios.
Although our method assumes object models to be exact and although running time grows quickly with the number of objects, we provide simple instructions to incorporate safety margins and to reduce compute time if needed.
\end{abstract}


\maketitle

\section{Introduction}
\IEEEPARstart{A}{ssessing}
the capacity of multi-object assemblies to withstand external forces without becoming unstable is crucial to enable robots to make informed decisions in real-world environments, where autonomy hinges on accurate and reliable world models.
Mobile robots efficiently traversing cluttered environments \cite{heins_2023_keep, flores_2013_time}, robotic manipulators planning grasps \cite{omata_rigid_2000, maeda_analysis_2007, bicchi_robotic_2000} or placing objects \cite{wang_dense_2021, srinivas_busboy_2023, jiang_2012_learning}, autonomous construction robots \cite{wermelinger_grasping_2021, johns_2023_framework}, and robots in many other contexts all need to answer the question: \textit{What forces can an assembly of objects in frictional contact withstand before becoming unstable?}
Answering this question correctly can support a robot into safely interacting with its surroundings and making informed decisions when handling and manipulating objects.

\begin{figure}[t]
    \centering
    \begin{overpic}[width=1\linewidth]{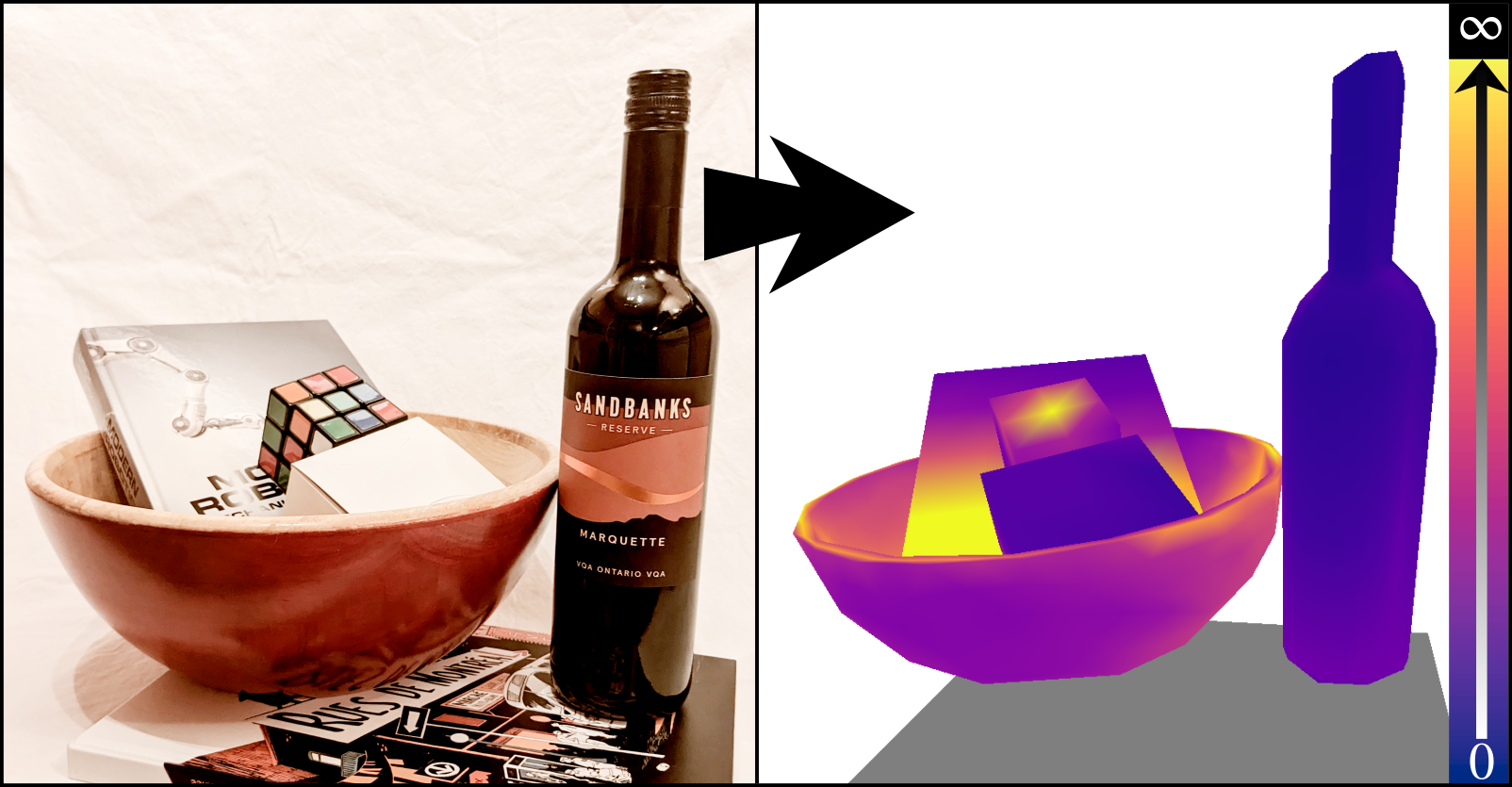}
        \put(18,43){\scalebox{1.5}{Scene}}
        \put(65,46){\scalebox{1.5}{Object}}
        \put(60,41){\scalebox{1.5}{Robustness}}
    \end{overpic}
    \caption{Using our method, a robot can assess the robustness of objects in complex scenes (lighter colors indicates higher robustness), allowing motion planners to avoid areas close to insecure objects (e.g., the wine bottle) and object placement planners to identify stable spots (e.g., the bottom of the book).}
    \label{fig:frontpage}
    \vspace{-3mm}
\end{figure}

This work contributes a novel method for evaluating the \textit{robustness} of rigid object assemblies, that is, their capacity to withstand external forces without moving, as pictured in \cref{fig:frontpage}.
We also describe how our method can be useful in industrial object handling applications, with experimental results demonstrating its practical effectiveness.
Due to the interplay of objects in contact, assessing the robustness of an assembly is complex and depends on the shapes, inertial parameters, frictional properties, and poses of all objects.
Although heuristic methods can suffice in select applications \cite{wang_2019_stable, wang_2010_heuristics}, a more general approach is needed to enable robots to handle diverse objects in a wide variety of scenarios.
Previous efforts have resorted to computationally expensive methods, such as dynamics simulations, to assess the robustness of assemblies \cite{lee_2023_object, chen_2021_planning, saxena_2023_planning}.
In contrast, this work proposes a method to compute an accurate solution for any rigid object assembly much more quickly than existing approaches.
Given the poses, inertial parameters, and friction coefficients of known objects in an assembly, our algorithm can assess the maximum force that can be sustained by the assembly before becoming unstable.
We validate our method against manually computed theoretical solutions and benchmark it against simulation-based and optimization-based approaches in terms of execution time and accuracy.
We believe that our solution to the aforementioned problem can provide a foundation for autonomous decision-making when interacting with objects in frictional contact.

\begin{figure}[b]
    \centering
    \vspace{-3mm}
    \setlength{\fboxsep}{0pt}%
    \setlength{\fboxrule}{0.5pt}%
    \fbox{\includegraphics[width=1\columnwidth-1pt]{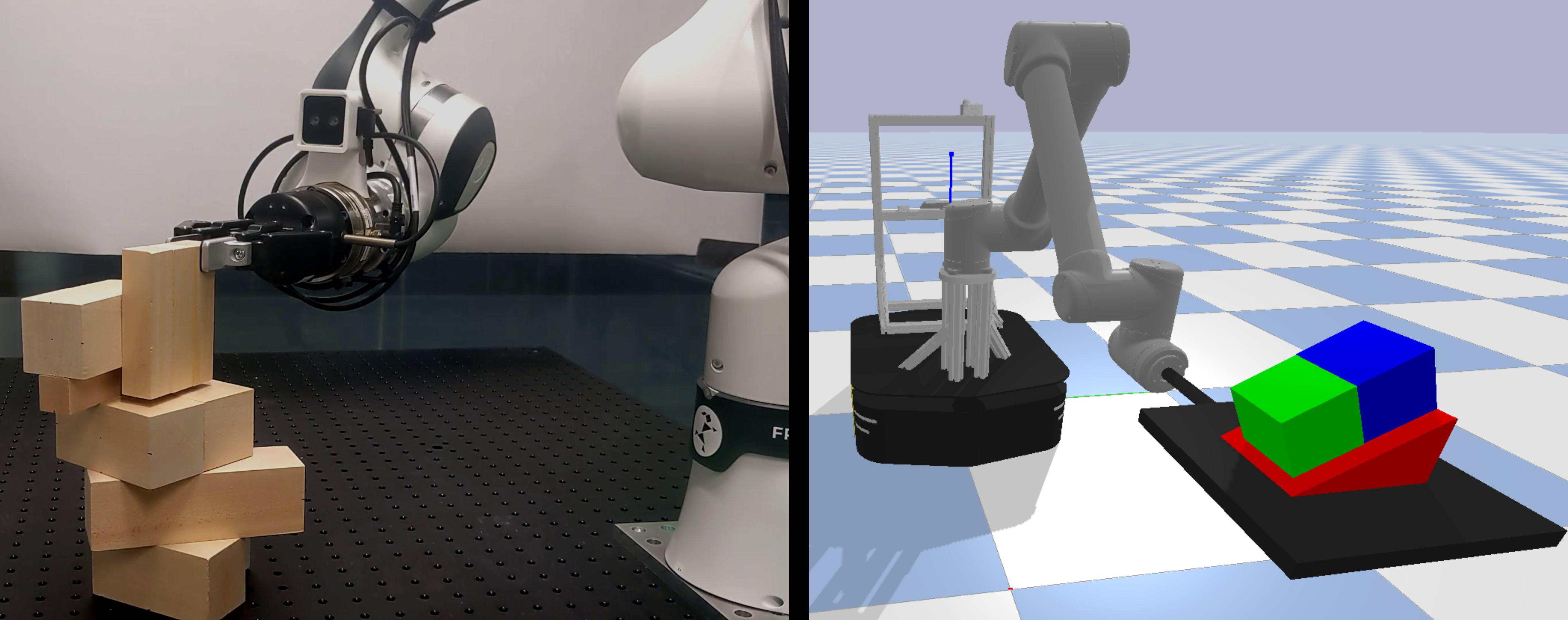}}
    \caption{Two key applications of our method: object placement planning \cite{nadeau_planning_2024} (left) and safe transportation \cite{heins_2023_keep} (right).}
    \label{fig:applications}
\end{figure}

To highlight potential applications of our method, we present three key use cases (two of which are illustrated in \cref{fig:applications}): stable object placement planning, safe assembly transportation, and disassembly planning.
In the first application, we show that an existing stable object placement planner \cite{nadeau_planning_2024} can produce better placements when relying on our proposed method.
In the second application, we demonstrate how our method can be used to determine the acceleration that a mobile manipulator should not exceed when transporting an assembly of objects.
Finally, we also expose how our method can be used to define disassembly sequences, enabling a robot manipulator to remove objects from an assembly without destabilizing it.
Overall, our proposed method aims at improving the autonomy of mobile manipulators in industrial object handling tasks involving disassembling, transporting, and placing objects safely.

The remainder of this paper is organized as follows.
In \cref{sec:related_work}, we highlight prior work on friction modeling, contact force resolution, and assembly stability assessment.
The optimization problem used to compute contact forces is detailed in \cref{sec:computing_reaction_forces}, and our proposed method for assessing the robustness of an assembly is presented in \cref{sec:computing_static_robustness}.
A theoretical validation of our method is provided in \cref{sec:validation}, where we also benchmark three existing methods against ground truth solutions.
In \cref{sec:applications}, we highlight three key applications of our method: stable placement planning, safe assembly transportation, and disassembly planning.
Finally, an outlook on sources of errors, advantages, and limitations of our work is provided in \cref{sec:discussion}.

\section{Related Work}
\label{sec:related_work}

\subsection{Dynamics of Rigid Objects in Frictional Contact}
\label{sec:rel_work_friction}

Assuming that discrete contact points constitute the interface between two objects, the onset of slipping---which leads to instability---is governed by local forces at discrete contact points. %
For rigid objects in the elastic regime, \cite{shigley_mechanical_1972} and \cite{sinha_contact_1990} conclude that local forces at contact points on an interface between two objects are cumulative.
They also show that the object deformation under load is such that the total potential energy of the system is minimized.
Furthermore, results from \cite{otsuki_systematic_2013} indicate that, although the local stress at contact points can be momentarily released, all contact points are maximally stressed at the onset of bulk slipping.
In \cite{bhushan_tribology_2013}, friction is described not as a material characteristic but as a systems response that is conditioned, in part, by (i) asperities in the contact surface (e.g., ploughing), (ii) adhesion between surfaces (e.g., chemical bonding), and (iii) lubrication of the contact surface (allowing a layer of atoms to slide relative to a deeper layer).
Although friction depends on factors including contact time and temperature, experiments in \cite{bhushan_contact_1998} suggest that the tangential force required to trigger slipping can be approximated reasonably well by the Coulomb friction model \cite{coulomb_theorie_1821} for rigid objects in the elastic regime.
Our proposed algorithm makes use of the findings from \cite{sinha_contact_1990} and \cite{otsuki_systematic_2013} to model force interactions for objects with multiple contact points.
Also, we base our approach on the Coulomb friction model to make it practical for common robot applications.

For an assembly to be stable, all objects must be under force equilibrium; and enforcing this condition necessitates computing forces at the contact points.
However, \cite{mattikalli_finding_1996} highlights that the equilibrium condition is, in general, insufficient to guarantee stability due to indeterminacies in the contact forces.
Furthermore, \cite{mason_mechanics_1986} indicates that since the deformation of an object (assumed to be rigid) is much smaller than the uncertainty in the object's shape, the location of contact points is effectively indeterminate in practice.
In this work, we avoid static indeterminacies by relaxing the assumption that objects are perfectly rigid, an approach that is coherent with the principle of virtual work \cite{sinha_contact_1990}.
The optimization problem we use to determine contact forces is also central to methods used in the computer-aided design of rigid structures \cite{whiting_structural_2012, kao_coupled_2022, yao_interactive_2017}, and is equivalent to the solution obtained with finite-element methods \cite{shin_reconciling_2016}.

Rigid-body dynamics for multi-object simulations is reviewed in \cite{stewart_rigid-body_2000, liu_2021_pbs}, and a survey of techniques and challenges related to grasping and fixturing is presented in \cite{bicchi_robotic_2000}.

\subsection{Stability Assessment}
\label{sec:rel_work_stability}

Assessing the stability of objects in contact is useful in a breadth of applications such ase grasp planning \cite{omata_rigid_2000, maeda_analysis_2007, bicchi_robotic_2000, wermelinger_grasping_2021}, automated fixturing \cite{sugar_metrics_2000}, object placement planning \cite{wang_dense_2021, srinivas_busboy_2023}, computer-aided design \cite{whiting_structural_2012, kao_coupled_2022}, object transportation \cite{heins_2023_keep, flores_2013_time}, and others.
While some methods focus on a single object (e.g., planning how to grasp or fixture a given object), the algorithm proposed in this work can be used to assess the stability of a multi-object assembly and compute its robustness to external forces.

Abstracting away from contact forces, \cite{mojtahedzadeh_support_2015} considers support relationships between objects to facilitate convex object placement planning.
In turn, support relationships are determined through the use of geometric heuristics in \cite{kartmann_extraction_2018} that do not consider inertial parameters or friction coefficients.
As a result, the applicability of these methods are limited to simple assemblies for which heuristics can be defined.
In contrast, our approach can be applied to any assembly of rigid objects.

Our method relies on a graph representation of the assembly, similar to the one used in \cite{lee_1999_disassembly, lee_1994_forceflownetwork} to plan disassembly sequences.
The graph representation in \cite{boneschanscher_subassembly_1988} uses edges to represent the set of forces that an object can exert onto another without making it move. 
However, since the set of forces that object \textit{A} can exert onto object \textit{B} is not equal to the set of forces that \textit{B} can exert onto \textit{A}, \cite{boneschanscher_subassembly_1988} is limited to assemblies that can be described with directed acyclic graphs (i.e., without loops).
In contrast, the nature of the graph used in our work makes our proposed algorithm applicable to any assembly of rigid objects.

The assembly robustness criterion proposed in \cite{maeda_new_2009} is defined as an optimization problem in which the magnitude of a given external wrench is maximized.
For each external wrench query, \cite{maeda_new_2009} tackles the full computation of contact forces and robustness assessment in a single optimization problem.
Similarly, \cite{chen_2021_planning} performs stability analysis by solving an optimization problem for a large number of random force disturbances, resulting in large running times due to their stability assessment scheme.
In contrast, our proposed approach decomposes the overall problem into sub-problems and use object shape information to improve efficiency. 

In \cite{nadeau_planning_2024}, an approximate robustness assessment method is used to find stable object placement poses much faster than with other methods.
Compared to the heuristic used in \cite{nadeau_planning_2024}, the method defined in this work assesses the assembly robustness more accurately, albeit at a higher computational cost.
In this work, we show that upgrading \cite{nadeau_planning_2024} with our method can improve the robustness of the placement poses found and reduce the planning time when dealing with complex scenes.

\section{Notation}
In this work, rigid transformations are denoted using the RIGID notation convention \cite{nadeau_rigid_2024} where $\Pos{a}{b}{c}$ is the position vector of $\CSys{a}$ with respect to $\CSys{b}$ as expressed in coordinate system $\CSys{c}$, and $\Rot{a}{b}$ is the rotation matrix expressing the orientation of $\CSys{a}$ with respect to $\CSys{b}$. 
We use $\UnsetConciseNotation \Pos{a}{b}{b} \equiv \Pos{a}{b} \SetConciseNotation$ for a more succinct notation.
We denote a unit-length vector $\Vector{v}$ by $\hat{\Vector{v}}$, the skew-symmetric operator by $\Skew{\Vector{u}}\Vector{v} = \Vector{u}\CrossP\Vector{v}$, and a $n\times n$ identity matrix by $\IdentityMatrix{n}$.
In this work, $\Vector{v}_{t}  \in \mathcal{S}^1$ and $\Vector{v}_{n} \in \Real$ respectively denote the tangential and normal components of $\Vector{v}$ where $\mathcal{S}^1$ is the unit circle in $\Real^2$.

\section{Computing contact forces}
\label{sec:computing_reaction_forces}
In this work, we make the following assumptions:
\begin{enumerate}
  \item objects are almost perfectly rigid, with local deformations at discrete contact points only;
  \item contact points are non-adhesive and deform elastically;
  \item the Coulomb friction model is in effect; and
  \item the shapes, inertial parameters, and coefficients of friction of all objects are known.
\end{enumerate}
While these assumptions are commonly used in robotics \cite{lynch_modern_2017, mason_mechanics_book_2001, boneschanscher_subassembly_1988}, an extensive study of their practical applicability concluded that the Coulomb friction model is reasonably accurate but its coefficients vary across contact area \cite{yu_more_2016}.
Expressed as an optimization problem, finding values for the contact forces in an assembly can yield no solutions, in which case the assembly is guaranteed to be unstable \cite{pang_stability_2000}.
Otherwise, a feasible set of contact forces stabilizing the assembly can be found.

\subsection{Objective Function}
\label{sec:objective_function}
For stiff materials, the force needed to slightly deform the contact points is governed by Young's modulus $\kappa$ with $\Norm{\RForce} = \kappa x$, where $x$ is the magnitude of the deformation.
The energy stored through the deformation is given by
\begin{equation}
    \label{eq:potential_deformation_energy}
    U = \int \Norm{\RForce} dx = \frac{\kappa x^2}{2} = \frac{\Norm{\RForce}^2}{2\kappa}
\end{equation}
and the \textit{principle of virtual work} dictates that an assembly will minimize this value \cite{sinha_contact_1990}.
Consequently, our objective function minimizes the sum of squared contact forces in the assembly.

\subsection{Equilibrium Constraints}
\label{sec:equilibrium_constraints}
A contact force $\RForce$ at a contact point can be decomposed into a frictional component $\RForce{t}$ and a normal component $\RForce{n}$, with the frictional component further decomposed into two orthogonal components $\RForce{u}$ and $\RForce{v}$.
A local reference frame $\CFrame{i} = \bbm \hat{\Vector{u}} & \hat{\Vector{v}} & \hat{\Vector{n}} \ebm $ can be defined, where $\hat{\Vector{u}}$, $\hat{\Vector{v}}$, and $\hat{\Vector{n}}$ are the unit vectors of the local frame with the inward surface normal given by $\hat{\Vector{n}}$.
The wrench exerted by $\RForce{i}{w}{w}$ on the object, and expressed in the assembly base reference frame $\CFrame{w}$, is 
\begin{equation}
    \label{eq:wrench_local_to_global}
    \Wrench{i}{w}{w} = 
    \underset{\Matrix{B}_i}{\underbrace{
            \bbm
            \IdentityMatrix{3}\\
            \Skew{\Pos{i}{w}{w}}
            \ebm
            \Rot{i}{w}
    }}
    \bbm \RForce{u} \\ \RForce{v} \\ \RForce{n} \ebm
    ,
\end{equation}
where $\Pos{i}{w}{w}$ and $\Rot{i}{w}$ are the position and orientation of $\CFrame{i}$ relative to $\CFrame{w}$, respectively.
The equilibrium of the $j$th object can be expressed as
\begin{equation}
  \sum_{k}^{\vert K_j\vert} \Matrix{B}_k \RForce{k} + \Wrench{g_j}{w}{w} = \Vector{0},
\end{equation}
where $\Wrench{g_j}{w}{w}$ denotes the wrench due to the object being accelerated (e.g., by gravity) and $K_j$ is the set of contact points on the object.

\subsection{Friction Magnitude Constraints}
\label{sec:friction_magnitude_constraints}

\begin{figure}[h]
    \centering
    \subfloat[]{
      \begin{overpic}[width=0.5\columnwidth]{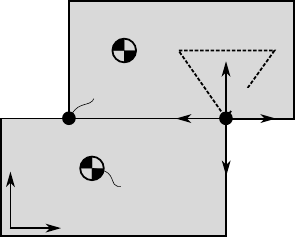}
        \put(5,7){$\CFrame{w}$}
        \put(38,15){$\Pos{c_1}{w}$}
        \put(28, 50){$\Pos{1}{w}$}
        \put(59,33){$-\CFrame{2}$}
        \put(79,43){$\CFrame{2}$}
        \put(71, 57){$\hat{n}$}
        \put(90, 34){$\hat{u}$}
      \end{overpic}
    }
    \subfloat[]{
      \begin{overpic}[width=0.4\columnwidth]{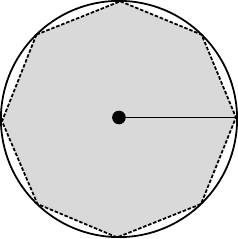}
        \put(62, 54){$\mu \Norm{\RForce{n}}$}
      \end{overpic}
    }
    \caption{(a) Illustration of the notation used in this work. (b) Octagonal approximation to the Coulomb circle.}
    \label{fig:coulomb}
\end{figure}

The Coulomb friction model \cite{coulomb_theorie_1821, mason_mechanics_1986} states that, for a contact point to avoid slipping,
\begin{equation}
    \label{eq:coulomb_friction}
    \Norm{\RForce{t}} \leq \mu \Norm{\RForce{n}},
\end{equation}
where $\mu \in \NonnegativeReal$ is the coefficient of friction, as shown in \cref{fig:coulomb}.

The Coulomb circle is commonly approximated \cite{mattikalli_finding_1996, stewart_implicit_2000} with an inscribed polygon, like a square or an octagon, which respectively cover about $65\%$ and $90\%$ of the circle.
In any case, the inequalities
\begin{equation}
    \Matrix{C} \bbm \RForce{u} & \RForce{v} & \RForce{n} \ebm \Transpose \leq \Vector{0}_{N\times 1},
\end{equation}
where $\Matrix{C}$ is a $N\times 3$ matrix whose $k$-th row is given by 
\begin{equation}
    \Matrix{C}\left[k,1:3\right] = \bbm \cos\frac{2\pi k}{N} & \sin\frac{2\pi k}{N} & -\mu \cos\frac{\pi}{N} \ebm,
\end{equation}
constrain the friction force to lie within an inscribed $N$-sided regular polygon.%

\subsection{Unilaterality Constraints}
\label{sec:unilaterality_constraints}
The assumption that contact points are non-adhesive can be enforced with
\begin{equation}
    \RForce{n} \geq 0
\end{equation}
assuming that $\hat{\Vector{n}}$, along which $\RForce{n}$ is directed, is the inward normal of the object surface at the contact point.

\subsection{Optimization Problem}
\label{sec:solving_forces_problem}
Given a set of $J$ objects, let $I$ be the set of all contact points across all objects and $K_j$ be the set of contact points on object $j \in J$.
The complete formulation of the optimization problem for computing contact forces in an assembly is
\begin{align}
    &\underset{\RForce_i}{\text{min}} \quad \sum_i^{\vert I \vert} \Norm{\RForce_i}^2\\
    &\text{subject to}\notag\\
    &\sum_{k}^{\vert K_j\vert} \Matrix{B}_k \RForce{k} + \Wrench{g_j}{w}{w} = \Vector{0} \hspace{7mm} \forall j \in J\\
    &\Matrix{C}_i \RForce_i \leq \Vector{0} \hspace{2.75cm} \forall i \in I\\
    &\RForce{i_n} \geq 0 \hspace{2.95cm} \forall i \in I
\end{align}
representing a quadratic program with linear constraints that can be efficiently solved with standard methods \cite{osqp}.

\section{Robustness Computation}
\label{sec:computing_static_robustness}
In this work, we define \textit{robustness} as the amount of force that can be applied in a given direction on a given point before relative motion occurs between objects in an assembly.
We formally define the robustness as
\begin{equation}
    \SR : \Real^3 \times \mathbb{S}^2 \rightarrow \NonnegativeReal
\end{equation}
which maps a pair $(\Pos{i}{w} \in \Real^3~,~\EForce[\hat]{i}{w} \in \mathbb{S}^2)$ to the maximum force magnitude $\SR \in \NonnegativeReal$ that can be exerted on point $\Pos{i}{w}$ in the direction $\EForce[\hat]{i}{w}$ before the onset of motion occurs, and where $\mathbb{S}^2$ is the unit sphere.

\subsection{Contact Interface Graph}
\label{sec:contact_interface_graph}
\begin{figure}[ht]
    \centering
    \subfloat[]{
        \includegraphics[width=0.5\columnwidth]{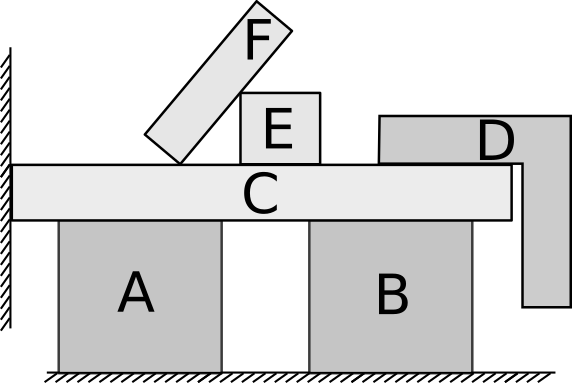}
    }
    \hspace{5mm}
    \subfloat[]{
        \includegraphics[width=0.3\columnwidth]{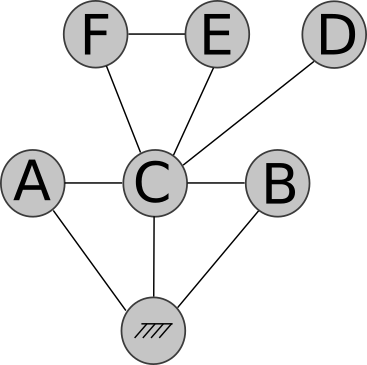}
    }
    \caption{(a) A structure composed of six non-fixed objects and two fixed objects (i.e., floor and wall). (b) The contact interface graph of the structure with the fixed nodes at the bottom.}
    \label{fig:cig}
\end{figure}

An arbitrary assembly of rigid objects can be described with a \textit{contact interface graph} (CIG), in which each node represents an object in the assembly and each edge represents a contact interface \cite[Chapter 37]{handbook_robotics_2008} between two objects, as shown in \cref{fig:cig}.
Describing assemblies with this type of graph can be traced back to \cite{lee_subassembly_1994}.

The information contained in the CIG can be stored as a list of interfaces, each defined by the two objects it connects, the locations of the contact points on the interface, and the forces at the contact points.
To simplify the graph, nodes representing fixed objects (e.g., a floor or a wall) can be merged into a single node marked as being \textit{fixed}.
Since no object can be free-floating, the resulting graph is guaranteed to be connected such that a path exists from any node to the fixed node.

\subsubsection{Modeling Instabilities as Graph Cuts}
\label{sec:modeling_instabilities_as_graph_cuts}
\begin{figure}
  \centering
  \subfloat[Translational relative motion]{
    \includegraphics[width=\columnwidth]{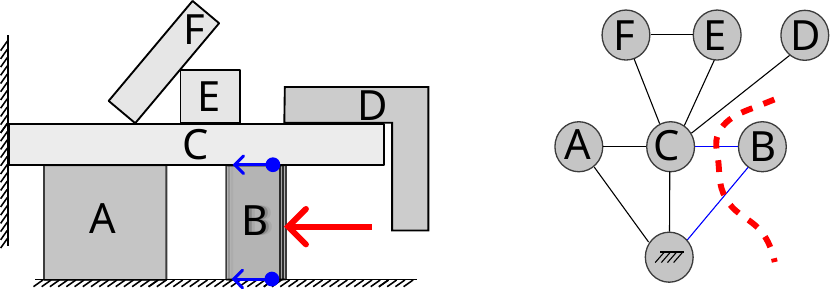}
  }\\
  \subfloat[Rotational relative motion]{
    \includegraphics[width=\columnwidth]{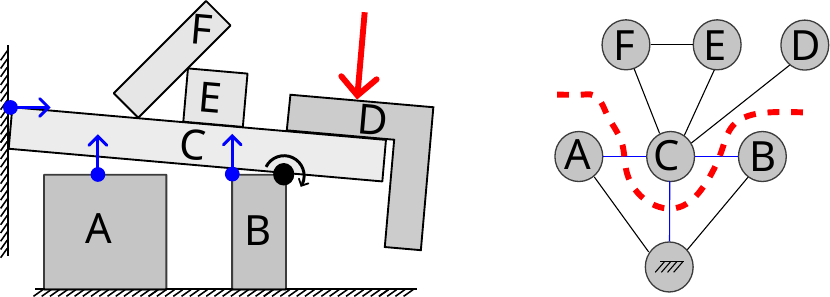}
  }
  \caption{Translational (a) and rotational (b) relative motions at the contact interfaces (small blue arrows) due to external forces (large red arrows). Both phenomena are modelled as a graph cut of the CIG, where edges cut represent contact interfaces at which relative motion happens.}
  \label{fig:graph_cut}
\end{figure}

At the onset of instability, a subset of objects in the assembly will move relative to the others.
This phenomenon can be modelled as a graph cut of the CIG, where edges cut represent contact interfaces at which relative motion between objects occurs.
The relative motion can be translational or rotational, as illustrated in \cref{fig:graph_cut}.
The former is associated with a slipping phenomenon while the latter is associated with a toppling phenomenon.
While slipping occurs when friction constraints are violated, toppling happens when an external force disturbs the equilibrium of the assembly.
The minimal force producing a given cut will either generate slipping or toppling, since a strictly greater force would be required to generate a toppling motion while objects are slipping.
Crucially, this enables dividing the overall problem into two separate sub-problems: computing the robustness to slipping, and computing the robustness to toppling. 

The overall robustness to a force exerted in direction $\EForce[\hat]{i}{w}$ on a point $\Pos{i}{w}$ is given by 
\begin{equation}
  \SR\left(\Pos{i}{w}, \EForce[\hat]{i}{w}\right) = \min\left(\SR_{slip}\left(\Pos{i}{w}, \EForce[\hat]{i}{w}\right), \SR_{top}\left(\Pos{i}{w}, \EForce[\hat]{i}{w}\right)\right) 
\end{equation}
where $\SR_{slip}(\cdot)$ and $\SR_{top}(\cdot)$ are the robustness to slipping and toppling, respectively.
A key part of our approach is to deal with slipping and toppling separately, as they are fundamentally different phenomena, before combining their results.

\subsection{Robustness to Slipping}
\label{sec:robustness_to_slipping}
In the following, we first determine the robustness to an external force when considering a single contact point.
Then, we define how the robustness of several contact points on a common interface are combined.
Finally, we explain how the robustness to slipping for a complete assembly can be obtained.

\subsubsection{At a Single Contact Point}
\begin{figure}
    \centering
    \subfloat[$d<0$, $\frac{\partial \CC}{\partial \Mag}(0)<0$]{
        \begin{overpic}[width=0.3\textwidth]{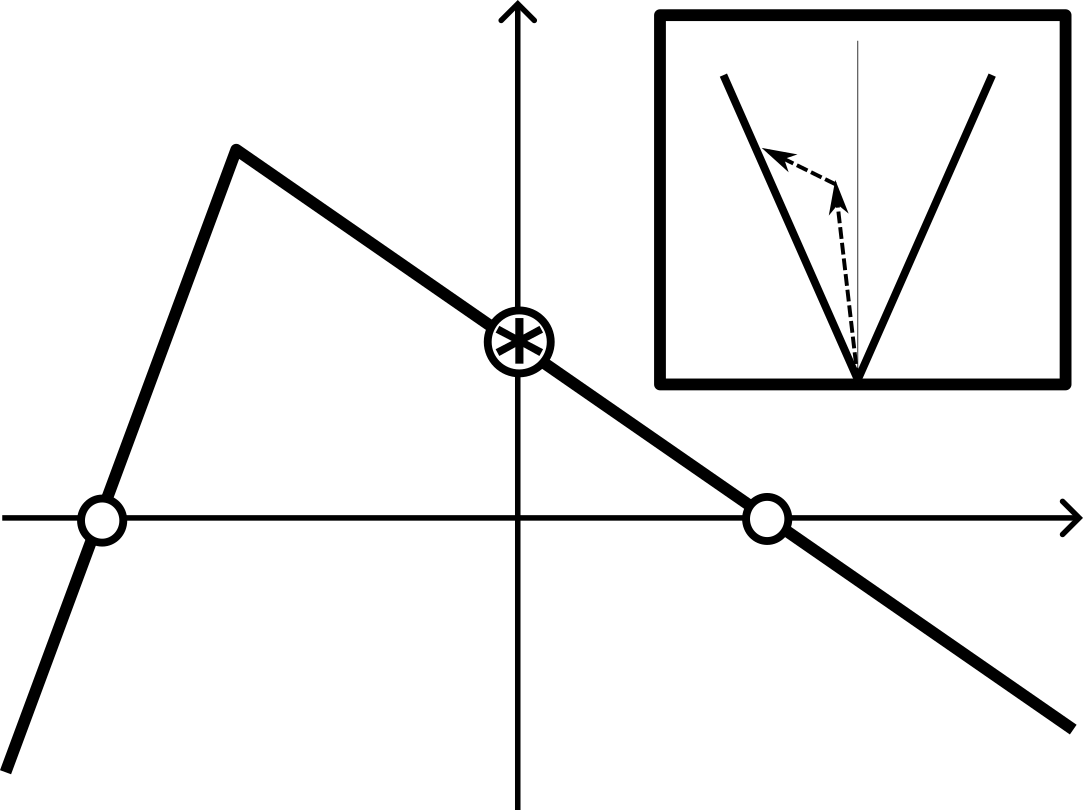}
            \put(50,72){$\CC(\Mag)$}
            \put(97,30){$\Mag$}
            \put(70,30){$\Mag_m$}
            \put(10,20){$\Mag_p$}
            \put(10,20){$\Mag_p$}
            \put(48,47){$\Mag^\star$}
        \end{overpic}
        \label{fig:msf_contact_conditions1}
    }
    \hspace{1cm}
    \subfloat[$d<0$, $\frac{\partial \CC}{\partial \Mag}(0)>0$]{
        \begin{overpic}[width=0.3\textwidth]{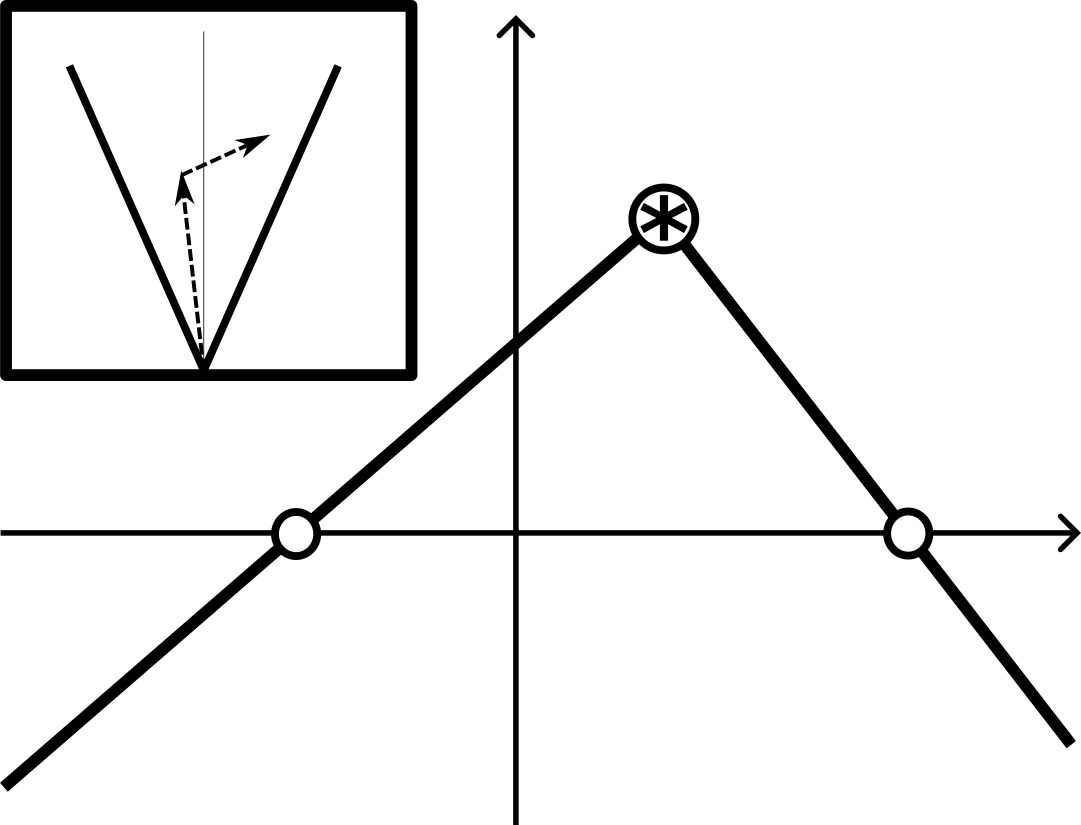}
            \put(50,72){$\CC(\Mag)$}
            \put(97,30){$\Mag$}
            \put(82,30){$\Mag_m$}
            \put(25,20){$\Mag_p$}
            \put(60,60){$\Mag^\star$}
        \end{overpic}
        \label{fig:msf_contact_conditions2}
    }\\
    \subfloat[$d>0$, $\frac{\partial \CC}{\partial \Mag}(0)>0$]{
        \begin{overpic}[width=0.3\textwidth]{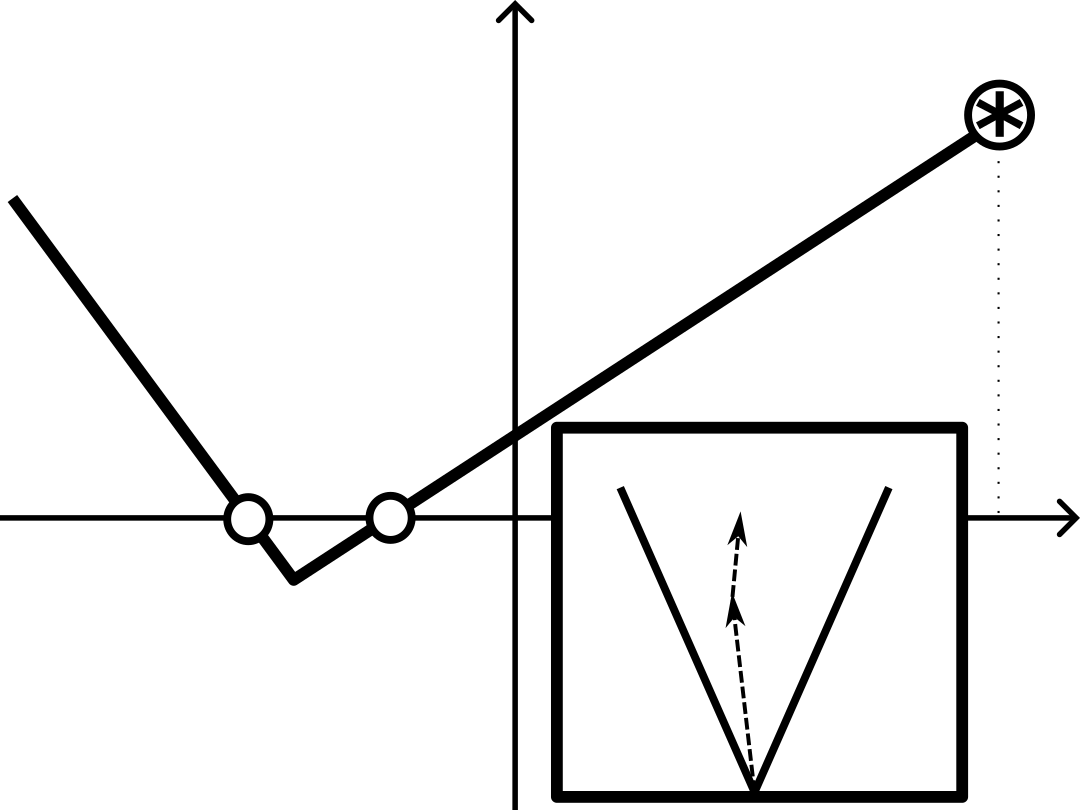}
            \put(50,72){$\CC(\Mag)$}
            \put(97,30){$\Mag$}
            \put(22,30){$\Mag_m$}
            \put(36,20){$\Mag_p$}
            \put(90,22){$\infty$}
            \put(91,68){$\Mag^\star$}
        \end{overpic}
        \label{fig:msf_contact_conditions3}
    }
    \hspace{1cm}
    \subfloat[$d>0$, $\frac{\partial \CC}{\partial \Mag}(0)<0$]{
        \begin{overpic}[width=0.3\textwidth]{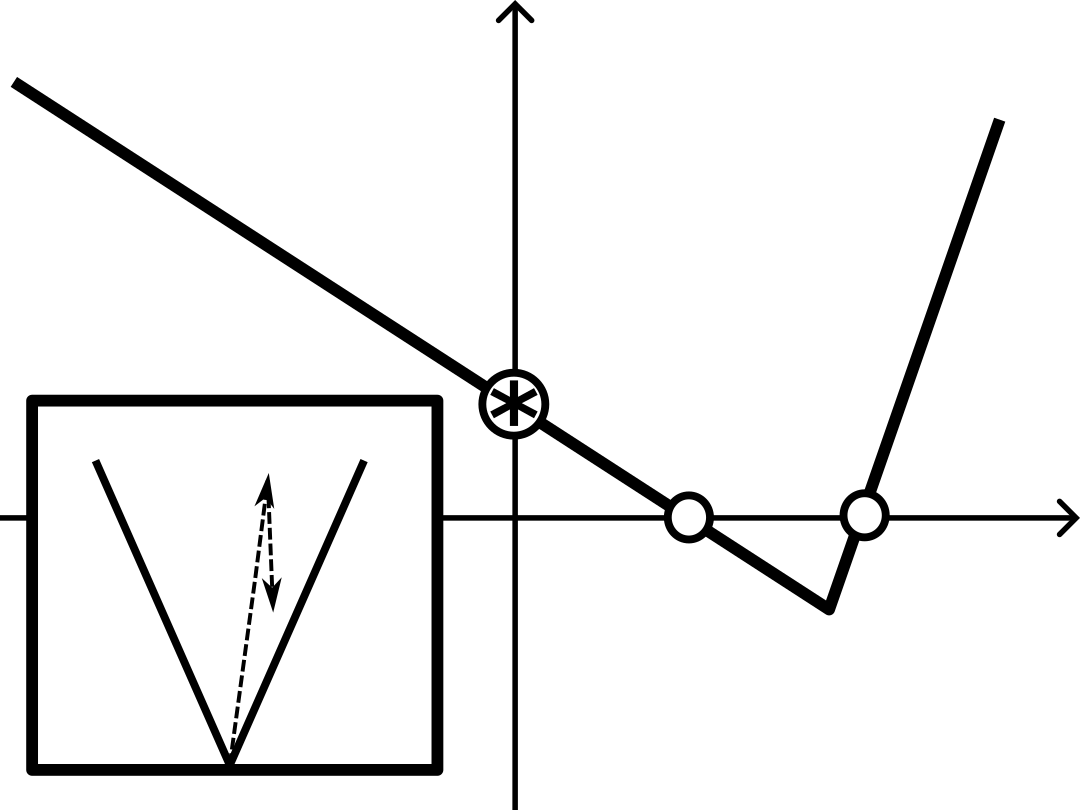}
            \put(50,72){$\CC(\Mag)$}
            \put(97,30){$\Mag$}
            \put(62,30){$\Mag_m$}
            \put(80,21){$\Mag_p$}
            \put(50,40){$\Mag^\star$}
        \end{overpic}
        \label{fig:msf_contact_conditions4}
    }
    \caption{The contact condition as a function of the external force magnitude. The roots of $\CC(\Mag)$ are $\Mag_m$ and $\Mag_p$ while $\Mag^\star$ is the magnitude maximizing $\CC$. Inset figures represent the friction cone with the external force starting at the tip of the contact force vector.}
    \label{fig:msf_contact_conditions}
\end{figure}

Let the contact force $\RForce$ at a given contact point be expressed in a local frame as $\RForce = \bbm \RForce{u} & \RForce{v} & \RForce{n} \ebm\Transpose$ where $\RForce{t} = \bbm \RForce{u} & \RForce{v} \ebm\Transpose \in \mathcal{S}^1$ is the tangential force and $\RForce{n}$ is the normal force, as pictured in \cref{fig:coulomb}.
Also, let an external force applied at the contact point be expressed in the local frame as $\EForce = \Mag\EForce[\hat] = \bbm \EForce{u} & \EForce{v} & \EForce{n} \ebm\Transpose$ where $\Mag \in \NonnegativeReal$ is the magnitude of the external force and $\EForce[\hat] \in \mathbb{S}^2$ is its direction.
We define the \textit{contact condition} at the point as
\begin{equation}
    \label{eq:cc_function_of_mag}
    \CC(\Mag) = \mu \Norm{\RForce{n} + \Mag\EForce[\hat]{n}} - \Norm{\RForce{t} + \Mag\EForce[\hat]{t}}
\end{equation}
that equals zero when
\begin{equation}
    \label{eq:single_contact_slipping_condition}
    \mu \Norm{\RForce{n} + \Mag\EForce[\hat]{n}} = \Norm{\bbm \RForce{u} + \Mag\EForce[\hat]{u} & \RForce{v} + \Mag\EForce[\hat]{v} \ebm},
\end{equation}
at which point the contact point is on the verge of slipping.
Intuitively, the contact condition is the amount of force that can be withstood in a given direction before slipping happens.
Equation \cref{eq:single_contact_slipping_condition} is quadratic in $\Mag$ with solutions given by
\begin{align}
    \label{eq:slipping_solution}
    s_m &= \frac{\bbm \RForce{u}&\RForce{v}&-\mu^2 \RForce{n}\ebm \EForce[\hat]-n}{d}~,~s_p = s_m + \frac{2n}{d},
\end{align}
where 
\begin{align}
    n = \Big( 
        \mu^2\left( \EForce[\hat]{n}\RForce{u}-\EForce[\hat]{u}\RForce{n} \right)^2
        &+ \mu^2\left( \EForce[\hat]{n}\RForce{v}-\EForce[\hat]{v}\RForce{n} \right)^2 \notag\\
        &- \left( \EForce[\hat]{v}\RForce{u}-\EForce[\hat]{u}\RForce{v} \right)^2 
        \Big)^\frac{1}{2}
\end{align}
and
\begin{align}
    d = \mu^2\Norm{\EForce[\hat]{n}}^2-\Norm{\EForce[\hat]{t}}^2.
\end{align}
The solution in \cref{eq:slipping_solution} can be shown to be equivalent to the geometric problem of finding intersection points between an external force vector and the friction cone at a contact point, as derived in \cite{nadeau_planning_2024}.
The curvature of the equation in \cref{eq:single_contact_slipping_condition} is given by $d$ such that
\begin{equation}
    \Mag_m = \frac{\Norm{\RForce{t}}^2-\mu^2\Norm{\RForce{n}}^2}{2\left(\mu^2\Norm{\EForce[\hat]{n}}\Norm{\RForce{n}}-\EForce[\hat]{t}\Tran\RForce{t}\right)} \hspace{4mm} \text{when} \hspace{4mm} d = 0
\end{equation}
is the solution to the linear equation obtained when $\mu^2\Norm{\EForce[\hat]{n}}^2=\Norm{\EForce[\hat]{t}}^2$.
Assuming that $\CC(0) \geq 0$ (i.e., the assembly is stable when no external force is applied), four situations can be considered depending on the sign of $d$ and on the sign of $\partial \CC / \partial s$ at $\Mag=0$.
The four situations are shown in \cref{fig:msf_contact_conditions}, with inset figures representing the friction cone with forces acting at the contact point.
When positive, the solution $\Mag_m$ represents the maximum amount of force that can be exerted at the contact point along $\EForce[\hat]$ before slipping occurs ($\Mag_p$ could equivalently be used for this purpose).
When $\Mag_m < 0$, an infinite amount of force can be exerted along $\EForce[\hat]$ without slip occurring.
Hence, we define the robustness of the kth contact point to slipping when $\EForce[\hat]{i}{w}$ is exerted on point $\Pos{i}{w}$ as
\begin{equation}
    \label{eq:cp_slipping_robustness}
    \SR_{k_{slip}}(\Pos{i}{w}~,~\EForce[\hat]{i}{w}) = 
    \begin{cases}
        \Mag_m  \quad&\text{if } \Mag_m \geq 0\\
        \infty  \quad&\text{if } \Mag_m < 0
    \end{cases}
    .
\end{equation}

The situation depicted in \cref{fig:msf_contact_conditions1} represents the case where any non-zero external force (starting from $\Mag=0$) will reduce $\CC(\Mag)$ (the contact condition).
In \cref{fig:msf_contact_conditions2}, increasing the magnitude of the external force will first improve the contact condition by bringing the total force closer to the axis of the friction cone, but increasing it further will monotonically decrease the contact condition.
In \cref{fig:msf_contact_conditions3}, the external force is inside the friction cone at the contact point, and any amount of force will improve the contact condition.
In \cref{fig:msf_contact_conditions4}, the external force is inside the cone that is opposed to the friction cone and increasing the magnitude of the external force will decrease the contact condition until $\CC(\Mag)=0$. %

\subsubsection{At a Contact Interface}
\label{sec:robustness_to_slipping_interface}
Assuming that objects obey the laws of linear elasticity at the contact points, the stress incurred at each contact point is cumulated to produce the total frictional force \cite{sinha_contact_1990, bhushan_contact_1998}.
Locally, the stress on an contact point can be momentarily released with another one taking over the load, but the gross effect is that the total frictional force is the sum of the frictional forces at each contact point \cite{otsuki_systematic_2013}.
This is consistent with the \textit{principle of least action}: the system will naturally minimize its energy by distributing the forces as uniformly as possible on contact points, each respecting its traction limit.
Consequently, the robustness of a contact interface $t$ to an external force is given by 
\begin{equation}
  \label{eq:interface_slipping_robustness}
  \SR_{t_{slip}}\left(\Pos{i}, \EForce[\hat]\right) = \sum_k^{I_t} \SR_{k_{slip}}\left(\Pos{i}, \EForce[\hat]\right), 
\end{equation}
where $I_t$ is the set of contact points on the interface.

\subsubsection{For a Complete Assembly}
\label{sec:robustness_to_slipping_assembly}
\begin{figure}
  \centering
  \subfloat[Interfaces in series]{
    \begin{overpic}[width=0.75\columnwidth]{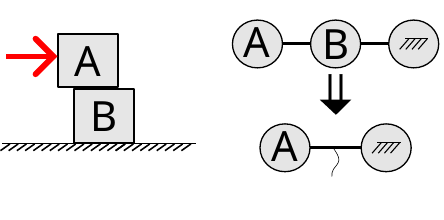}
      \put(63.5,43){$\SR_1$}
      \put(81.3,43){$\SR_2$}
      \put(61,5){$\min\left(\SR_1,\SR_2\right)$}
    \end{overpic}\label{fig:equiv_slip_rob_serial}
  }\\
  \subfloat[Interfaces in parallel]{
    \begin{overpic}[width=0.75\columnwidth]{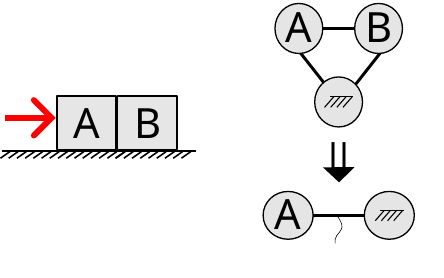}
      \put(66.1,4.2){$\SR_1+\SR_2$}
      \put(65,45){$\SR_1$}
      \put(85,45.5){$\SR_2$}
      \put(73.8,58){$\infty$}
    \end{overpic}\label{fig:equiv_slip_rob_parallel}
  }
  \caption{Combining interfaces acting (a) in series, and (b) in parallel to resist an external force (red arrow). The individual interface robustnesses $\SR_1$ and $\SR_2$ are combined through $\mathrm{min}\left(\cdot, \cdot\right)$ and $+\left(\cdot, \cdot\right)$ operators to combine interfaces in series and in parallel, respectively. }
  \label{fig:equiv_slip_rob}
\end{figure}

In general, an object will be in contact with several other objects through contact interfaces that have different maximal frictional force (i.e., traction limit).
For the reasons outlined in \cref{sec:robustness_to_slipping_interface}, we state the following assertion:
\begin{assertion}
\label{hyp:robustness_to_slipping_parallel_contacts}
The slipping robustness of several interfaces acting in parallel is given by the sum of the individual interface robustnesses.
\end{assertion}
An example of this situation is shown in \cref{fig:equiv_slip_rob_parallel}, where the total slipping robustness is given by the sum of the two interface robustnesses acting in parallel.

Furthermore, when two objects are in contact, the principle of action-reaction (i.e., Newton’s third law) implies that the force acting on an interface is exerted onto both objects (with opposite directions).
Hence, the maximum force that can be exerted at the interface by either object is given by the minimum traction of the two objects.
In other words, in regards to slipping, a chain of objects will only be as strong as the weakest interface along the chain.
This leads to the following assertion:
\begin{assertion}
\label{hyp:robustness_to_slipping_series_contacts}
The slipping robustness of several interfaces acting in series is given by the minimum of the interface robustnesses along the series.
\end{assertion}
An example of this situation is shown in \cref{fig:equiv_slip_rob_serial}, where the total slipping robustness is given by the minimum of the two interface robustnesses acting in series.

\begin{figure}
  \centering
  \begin{overpic}[width=1\columnwidth]{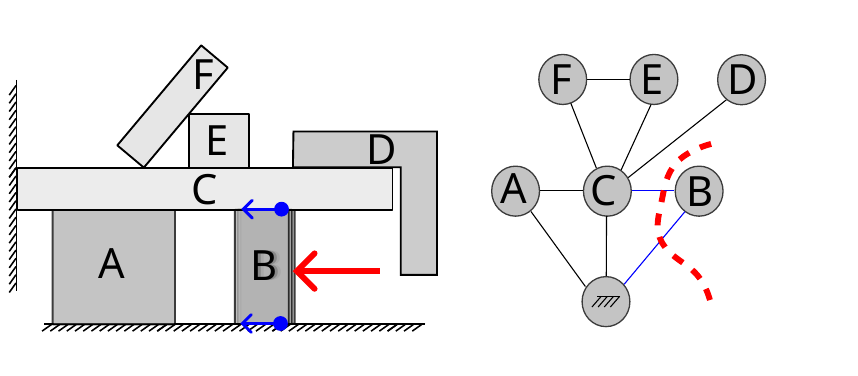}
    \put(85,20.5){\scalebox{0.8}{Source}}
    \put(68,3){\scalebox{0.8}{Sink}}
    \put(69,35){\scalebox{0.8}{$\infty$}}
    \put(65,26){\scalebox{0.8}{$2$}}
    \put(72,27){\scalebox{0.8}{$2$}}
    \put(80,30){\scalebox{0.8}{$3$}}
    \put(65,18.5){\scalebox{0.8}{$5$}}
    \put(74,18.5){\scalebox{0.8}{$6$}}
    \put(75,11){\scalebox{0.8}{$8$}}
    \put(62,13){\scalebox{0.8}{$9$}}
    \put(69.1,15){\scalebox{0.8}{$\infty$}}
  \end{overpic}
  \caption{The subset of objects that would slip under the exertion of the external force (large arrow) is given by the minimum cut (dashed line) of the CIG. The robustness of the assembly to the external force applied on the \textit{source} object is given by the maximum flow going into the \textit{sink}. Edge \textit{capacity} is given by the associated interface robustness.}
  \label{fig:slipping_max_flow}
\end{figure}

Given \cref{hyp:robustness_to_slipping_parallel_contacts} and \cref{hyp:robustness_to_slipping_series_contacts}, the problem of determining the slipping robustness of a target object in an assembly can be cast as a \textit{maximum flow} problem \cite{hillier_OR_2015}.
This is accomplished by defining a force flow network $G = (\mathcal{V}, \mathcal{E})$ where $\mathcal{V}$ is the set of nodes in the CIG and $\mathcal{E}$ is the set of edges.
In $G$,
\begin{enumerate}
  \item the node associated with the object onto which the external force is exerted is labeled as the \textit{source};
  \item the node associated with the fixed object is labeled as the \textit{sink}; and
  \item each edge is given a \textit{capacity} equal to the corresponding interface's robustness, as computed with \cref{eq:interface_slipping_robustness}.
\end{enumerate}
The graph cut minimizing the total capacity of the edges cut delineates the subset of objects that would slip under the exertion of the external force, as shown in \cref{fig:slipping_max_flow}.
Moreover, the slipping robustness of the assembly to the external force is given by the \textit{maximum flow} that can be routed from the source to the sink.
Provided that a maximum flow exists, the slipping robustness of the assembly is given by
\begin{equation}
  \label{eq:slipping_max_flow}
  \SR_{slip}(\Pos{i}, \EForce[\hat]) = \sum \rm{flow}\left(e\left[v, \rm{sink}\right]\right) \hspace{5mm} \forall~v \in \mathcal{V},
\end{equation}
where $e\left[v, \rm{sink}\right]$ represents the edge from node v to the sink if it exists, and $\rm{flow}(e)$ is the flow on edge $e$.
Standard algorithms, whose worst-came running time is $O(|J|^3)$ (cubic in the number of objects) or better, can be used to compute the maximum flow and minimum cut simultaneously \cite{hillier_OR_2015}.
This type of network has been used in \cite{lee_1994_forceflownetwork, lee_1999_disassembly} to study how clusters of parts in fixtured assemblies can be disassembled.

\subsection{Robustness to Toppling}
\label{sec:robustness_to_toppling}
While slipping occurs when an external force exceeds the traction of contact points in the assembly, toppling arises when the equilibrium of the assembly becomes impossible.
In the following, we provide an algorithm to compute the toppling robustness of an assembly to external forces, that is, the maximum permissible magnitude of an external force applied at a specific point before a subset of objects in the assembly rotate about an axis.
We start by defining feasible CIG cuts, and then determine axes about which motion can take place.
The set of graph cuts and toppling axes is finally used to compute the robustness of the assembly to toppling. 

\subsubsection{Defining Feasible Graph Cuts}
\label{sec:defining_feasible_graph_cuts}
\begin{figure}
  \centering
  \subfloat[Cuts of the CIG]{
    \begin{overpic}[width=1\columnwidth]{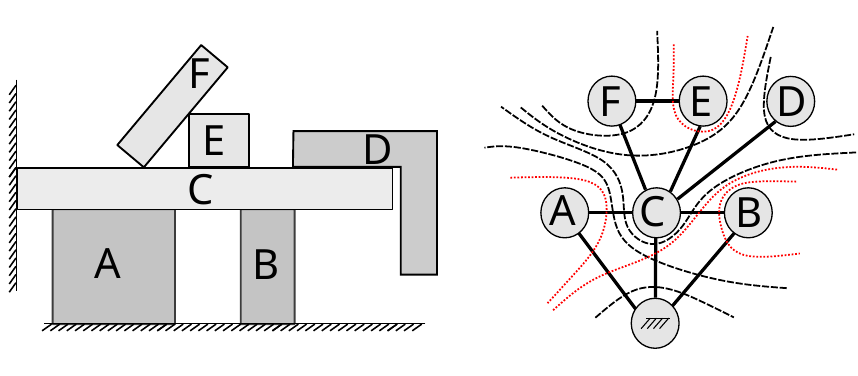}
    \end{overpic}\label{fig:toppling_all_cuts}
  }\\
  \subfloat[Toppling Axes]{
    \begin{overpic}[width=1\columnwidth]{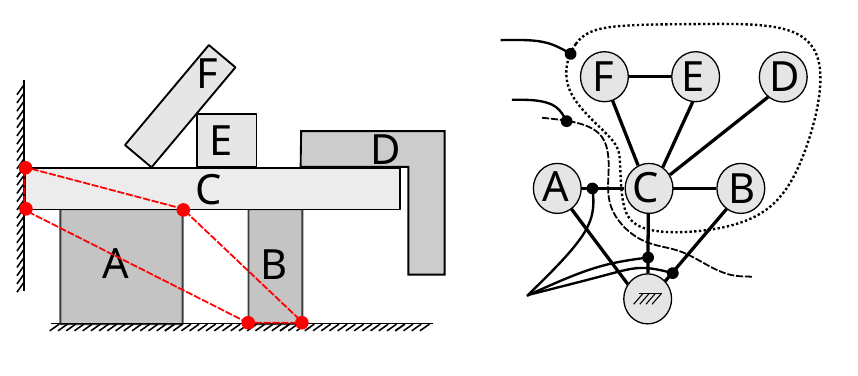}
      \put(55,38){\scalebox{0.8}{$\mathcal{G}_c$}}
      \put(57,31.5){\scalebox{0.8}{$c$}}
      \put(59,8){\scalebox{0.8}{$\mathcal{E}_c$}}
    \end{overpic}\label{fig:toppling_axes}
  }
  \caption{(a) Graph cuts of an assembly's CIG. Red dotted lines indicate infeasible cuts that isolate a form-closed object subset. (b) The edges of the convex hull over contact points (left side) located on interfaces cut (right side) define potential toppling axes (dots indicate through-page axes).}
\end{figure}

Similar to slipping, toppling can be modelled as a graph cut of the CIG, with the cut separating the objects that topple from those that remain stationary.
Any valid cut separates the CIG into two sub-graphs, one containing the fixed object and the other containing the toppling objects.
Some cuts are considered infeasible as they isolate a form-closed object subset \cite{bicchi_closure_1995} that cannot be toppled, as shown in \cref{fig:toppling_all_cuts}.
We define a \textit{feasible cut} as a cut separating the fixed object from other objects that, together, are not form-closed.
We denote the set of feasible cuts as $\mathcal{C}$ and the sub-graph (not containing the fixed node) produced by a cut $c \in \mathcal{C}$ as $\mathcal{G}_c$.
The set of edges cut by $c \in \mathcal{C}$ is denoted as $\mathcal{E}_c$ and the set of interfaces cut by $c$ is denoted as $\mathcal{T}_c$, as illustrated in \cref{fig:toppling_axes}.

\subsubsection{Defining Toppling Axes}
\label{sec:defining_toppling_axes}
For a given cut $c \in \mathcal{C}$, the edges of the convex hull over contact points $\left\{i \in I_t ~ \forall ~ t \in \mathcal{T}_c\right\}$ define potential toppling axes, as shown in \cref{fig:toppling_axes}.
In the particular case where all $t \in \mathcal{T}_c$ are parallel, an additional axis is defined normal to the contact interfaces and passing through the centre of friction \cite{mason_mechanics_1986}.
We denote the set of all potential toppling axes as $\mathcal{A}_c$.

\subsubsection{Defining Toppling Torque}
\label{sec:defining_toppling_torque}
For the ith contact point, we define the moment vector about toppling axis $a \in \mathcal{A}_c$ as 
\begin{equation}
  \label{eq:contact_point_moment}
  \VVec{i}{a} = \Pos{i}{a_s} \CrossP \UnitVec{a},
\end{equation}
where $\Pos{i}{a_s}$ is the position of the contact point relative to the toppling axis origin.
The countering moment exerted by the contact point about the toppling axis is given by
\begin{equation}
  \label{eq:cp_counteracting_force}
  {}_a \tau_i
  = \begin{cases}
    \left(\SR_{k_{slip}}(\Pos{i}{w},\EForce[\hat])\EForce[\hat]+\RForce{i}{w}\right)\DotP\VVec{i}{a}  \hspace{1.5mm} \text{if} \hspace{0mm} \VVec{i}{a}\DotP\UnitVec{n}_i = 0,\\
    \infty \hspace{40.2mm} \text{if} \hspace{0mm} \VVec{i}{a}\DotP\UnitVec{n}_i < 0,\\
    \RForce{i}{w}\DotP\VVec{i}{a} \hspace{31.2mm}\text{if} \hspace{0mm} \VVec{i}{a}\DotP\UnitVec{n}_i > 0,
  \end{cases}
  \hspace{-2.5mm}
\end{equation}
where $\SR_{k_{slip}}(\Pos{i}{w},\EForce[\hat])$ is computed from \cref{eq:cp_slipping_robustness}, $\UnitVec{n}_i$ is the surface normal at the contact point, and $\RForce{i}{w}$ is the contact force.
When $\VVec{i}{a}\DotP\UnitVec{n}_i = 0$, the toppling axis is perpendicular to the contact interfaces and friction forces are counteracting the toppling motion, while when $\VVec{i}{a}\DotP\UnitVec{n}_i < 0$, the contact force kinematically prevents toppling about the axis.
Otherwise, the contact force contributes to the equilibrium about the toppling axis and is used to compute the toppling torque.

\begin{figure}
  \centering
  \subfloat[Infeasible toppling]{
    \begin{overpic}[width=1\columnwidth]{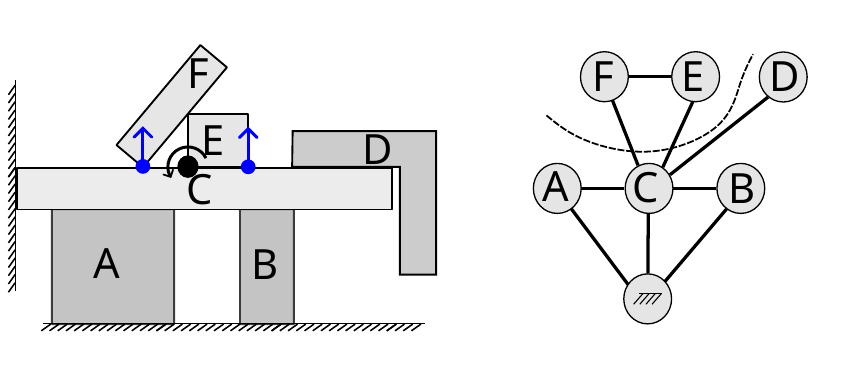}
      \put(69,28){\scalebox{0.8}{$\infty$}}
      \put(77,28){\scalebox{0.8}{$2$}}
    \end{overpic}\label{fig:toppling_invalid_axis}
  }\\
  \subfloat[Feasible toppling]{
    \begin{overpic}[width=1\columnwidth]{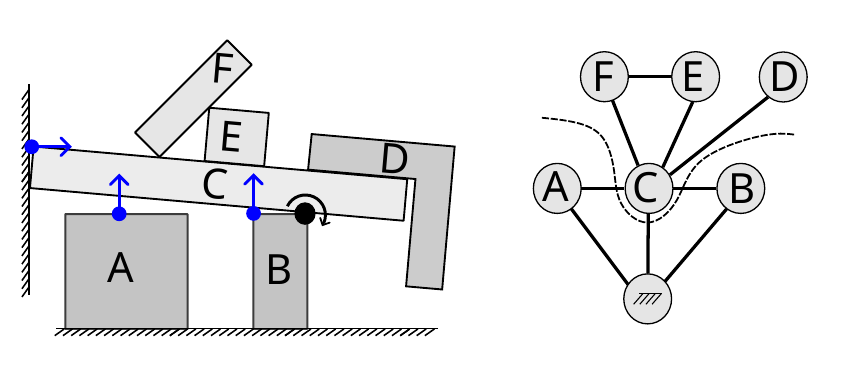}
      \put(68,19){\scalebox{0.8}{$10$}}
      \put(73.5,14){\scalebox{0.8}{$0$}}
      \put(81,19){\scalebox{0.8}{$5$}}
    \end{overpic}\label{fig:toppling_valid_axis}
  }
  \caption{(a) A toppling axis (black dot) deemed infeasible due to one of the edges cut contributing an infinite countering moment (right side). (b) The total toppling torque of edges cut (here equals to 15) is the torque required to produce toppling about the axis.}
\end{figure}

For a given cut $c \in \mathcal{C}$ and a given toppling axis $a \in \mathcal{A}_c$, the total toppling torque exerted by contact points in $I_t$ about $a$ is given by
\begin{equation}
    \label{eq:interface_torque}
    {}_a \tau_{I_t}
    = \sum_{i \in I_t} {}_a \tau_i, %
\end{equation}
where $I_t$ is the set of contact points on the interface $t \in \mathcal{T}_c$.
The total toppling torque exerted about an axis $a \in \mathcal{A}_c$ is given by 
\begin{equation}
    \label{eq:cut_torque}
    {}_a \tau_c
    = \sum_{t \in \mathcal{T}_c} {}_a \tau_{I_t}. %
\end{equation}
A toppling motion is deemed infeasible if ${}_a \tau_c$ is infinite, as is the case in \cref{fig:toppling_invalid_axis}.
For feasible axes, each edge's toppling torque indicates the contribution of contact forces at the edge's interface in counteracting toppling, as shown in \cref{fig:toppling_valid_axis}.

\subsubsection{Defining Toppling Robustness}
The toppling robustness for a set of point-direction tuples $\mathcal{P} = \left\{\left(\Pos{i}, \EForce[\hat]{i}\right)\right\}$ is computed  with \cref{algo:toppling_robustness}.
The initialization involves setting the toppling robustness of each element of $\mathcal{P}$ to infinity, and in determining the feasible cut set $\mathcal{C}$ as described in \cref{sec:defining_feasible_graph_cuts}.
For each cut $c \in \mathcal{C}$, a set of toppling axes $\mathcal{A}_c$ over the interfaces that are cut is determined following \cref{sec:defining_toppling_axes}.
For each toppling axis $a \in \mathcal{A}_c$, the toppling torque ${}_a \tau_c$ is computed with \cref{eq:cut_torque} and the toppling robustness of every point in $\mathcal{P}$ that is also in $\mathcal{G}_c$ is updated with 
\begin{equation}
    \label{eq:toppling_robustness}
    \SR_{top}\left(\Pos{i}, \EForce[\hat]{i}\right) = \min\left\{\SR_{top}\left(\Pos{i}, \EForce[\hat]{i}\right), \frac{{}_a \tau_c}{ \left(\Pos{i}{a_s}{w}\CrossP\EForce[\hat]{i}{w}\right)\DotP\UnitVec{a}}\right\}.
\end{equation}
The equation in \cref{eq:toppling_robustness} keeps the minimal toppling robustness, since the onset of instability will be triggered by the weakest force producing toppling.
\begin{algorithm}
    \caption{Computing Toppling Robustness}
    \label{algo:toppling_robustness}
    \DontPrintSemicolon
    \SetKwInOut{Input}{input}\SetKwInOut{Output}{output}
    \Input{CIG $G$, set of tuples $\mathcal{P} = \left\{\left(\Pos{i}, \EForce[\hat]{i}\right)\right\}$}
    \Output{toppling robustness $\SR_{top}\left(\Pos{i}, \EForce[\hat]{i}\right)~\forall~i \in \mathcal{P}$}
    1. Initialize $\SR_{top}\left(\Pos{i}, \EForce[\hat]{i}\right) \gets \infty~\forall~i \in \mathcal{P}$\;
    2. Determine the feasible cut set $\mathcal{C}$\;
    \ForEach{$c \in \mathcal{C}$}{
        \ForEach{$a \in \mathcal{A}_c$}{
          3. Compute ${}_a \tau_c$ with \cref{eq:cut_torque}\;
          4. $\forall i \in \mathcal{P} \cap \mathcal{G}_c$ :  Update $\SR_{top}\left(\Pos{i}, \EForce[\hat]{i}\right)$ with \cref{eq:toppling_robustness}. %
        }
    }
\end{algorithm}

The worst-case running time of \cref{algo:toppling_robustness} is $O\left(2^{|J|}\right)$, exponential in the number of objects in the assembly and proportional to the maximum number of cuts in the CIG.
Although the average running time can be much lower in practice, the theoretical worst case requires considering axes between every object combination.

\section{Validation and Benchmark Experiments}
\label{sec:validation}
\begin{figure}[b]
  \centering
  \begin{overpic}[width=1\columnwidth]{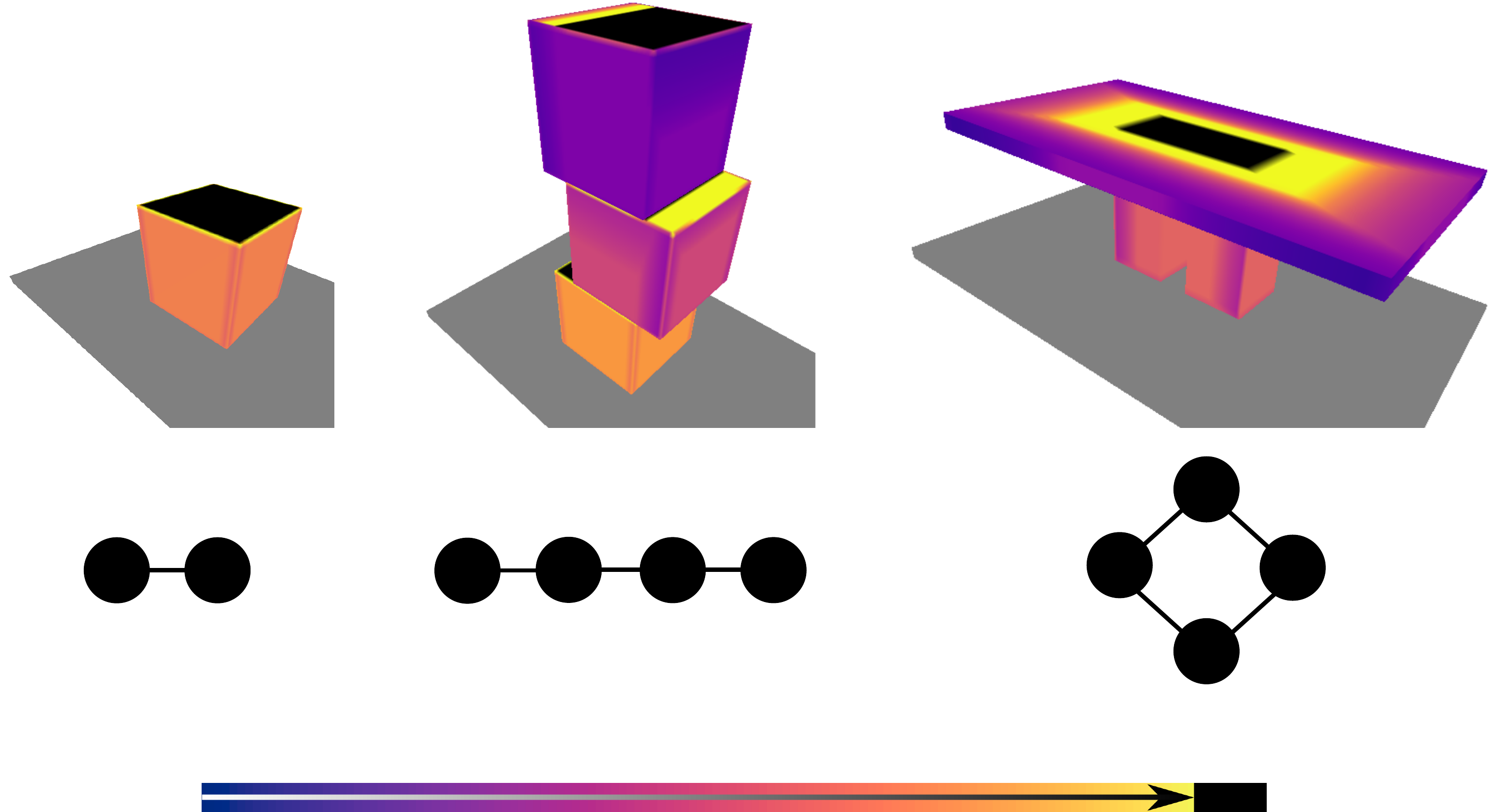}
    \put(11,-0.1){$0$}
    \put(80.5,0.1){{\color{white}$\infty$}}
    \put(6,5){\footnotesize (a) Cube}
    \put(36,5){(\footnotesize b) Stack}
    \put(75,5){(\footnotesize c) Table}
  \end{overpic}
  \caption{Assemblies used to benchmark the accuracy of robustness assessment methods with their respective CIG. The color of each surface point indicates the robustness to a normal force at that point according to the scale (bottom) with black indicating infinity.}
  \label{fig:validation_scenes}
\end{figure}

In the following, we compare the result of our method to three other methods and to manually-computed ground truth.
The comparison is done on three simple assemblies, shown in \cref{fig:validation_scenes}, for which it is feasible to manually compute the robustness of the assembly to external forces.
The first assembly is a single cube resting on a plane, the second assembly is a serial chain of three stacked cubes, and the third assembly is a slab resting on two legs and forming a parallel kinematic structure.
For visualization purposes, the robustness of the assemblies is computed for forces applied normally to the surface of the assembly.

\subsection{Benchmarked Methods}
\label{sec:validation_benchmark_methods}
We benchmark our algorithm against a method making use of commonly available simulators, a method iteratively solving optimization problems, and an approximate robustness heuristic.

\subsubsection{Simulation-based}
\label{sec:validation_benchmark_simulation}
Off-the-shelf dynamics simulators have been used to get a sense of an assembly robustness and to verify stability \cite{lee_2023_object, sun_stackgen_2024, yoneda_6dof_2023}.
These physics simulators can be used with general rigid body assemblies and are simple to set up.
To determine the robustness of an assembly to an external force applied at a given point, a binary search for the maximum force magnitude that can be sustained by the assembly before motion occurring is performed, as described in \cref{algo:simulator_robustness}.
To do so, initial searching bounds are set (i.e., [0, 100] Newtons) and a force whose magnitude is in the centre of the bounds is iteratively applied to the assembly.
Following the force simulation, assembly motion is determined by computing its kinetic energy and comparing it to a threshold value (i.e., 10$^{\text{-4}}$ Joules).
Whether the assembly moved or not dictates if the upper or lower bound is updated to the force magnitude.
The process is repeated a number of times (i.e., 100) before returning the centre of the bounds as the robustness of the assembly to the external force.

In our experiments, we used the PyBullet simulator \cite{coumans_2021} with the full Coulomb friction cone and with no force restitution to avoid bouncing effects.
The robustness search is done over 100 iterations, with 0--100 Newtons initial bounds, and a motion threshold of 10$^{\text{-4}}$ Joules.
\begin{algorithm}
    \caption{Simulator-based Robustness Search}
    \label{algo:simulator_robustness}
    \DontPrintSemicolon
    \SetKwInOut{Input}{input}\SetKwInOut{Output}{output}
    \Input{Query tuple $\left(\Pos{i}, \EForce[\hat]{i}\right)$}
    \Output{Robustness $\SR\left(\Pos{i}, \EForce[\hat]{i}\right)$}
    Initialize bounds $B_l,B_u \gets [0,100]$\;
    \For{$max\_iterations$}{
      $\Mag \gets (B_l + B_u)/2$\;
      Apply $\Mag\EForce[\hat]{i}$ at $\Pos{i}$ and simulate one time step\;
      Compute kinetic energy $E_{kin}$\;
      \eIf{$E_{kin} > threshold$}{
        $B_u \gets \Mag$\;
      }{
        $B_l \gets \Mag$\;
      }
    }
    $\SR\left(\Pos{i}, \EForce[\hat]{i}\right) \gets (B_l + B_u)/2$\;
\end{algorithm}

\subsubsection{Optimization-based}
\label{sec:validation_benchmark_optimization}
In \cite{maeda_new_2009, chen_2021_planning}, the optimization problem in \cref{sec:solving_forces_problem} is slightly modified into \cref{eq:opt_based_robustness_start}--\cref{eq:opt_based_robustness_end} to yield the robustness of an assembly to an external force.
The objective function of the original optimization problem is changed to one of maximizing the magnitude $\Mag$ of an external force applied at a given point $\Pos$ in the assembly.
The equilibrium constraint in \cref{eq:opt_based_robustness_equilibrium_constraint} for the object onto which the force is applied is modified to include the external force, while other constraints remain unchanged.
The resulting linear program
\begin{align}
  &\underset{\RForce_i,\Mag}{\text{max}} \quad \Mag\label{eq:opt_based_robustness_start}\\
    &\text{subject to}\notag\\
    &\Wrench{j} = \begin{cases}
      \Wrench{g_j}{w}{w} + \Mag\Matrix{B}_e\EForce[\hat] & \text{if } \Pos \text{ on object } j\\
      \Wrench{g_j}{w}{w} & \text{otherwise}
    \end{cases} \hspace{9mm} \forall j \in J\\
    &\sum_{k}^{\vert K_j\vert} \Matrix{B}_k \RForce{k} + \Wrench{j}  = \Vector{0} \hspace{3.8cm} \forall j \in J \label{eq:opt_based_robustness_equilibrium_constraint}\\
    &\Matrix{C}_i \RForce_i \leq 0 \hspace{5.5cm} \forall i \in I\\
    &\RForce{i_n} \geq 0 \hspace{5.7cm} \forall i \in I\label{eq:opt_based_robustness_end}
\end{align}
is solved for each query $\SR\left(\Pos, \EForce[\hat]\right)$.

To solve \cref{eq:opt_based_robustness_start}--\cref{eq:opt_based_robustness_end} in our experiments, we run the IPOPT solver \cite{wachter_implementation_2006} for a maximum of 1,000 iterations. 
If an optimal solution to the problem in \cref{eq:opt_based_robustness_start}--\cref{eq:opt_based_robustness_end} is found, we set $\SR\left(\Pos, \EForce[\hat]\right) = s$.
When the solver determines that the problem is infeasible, we set $\SR\left(\Pos, \EForce[\hat]\right) = 0$, and when the solver determines that the problem is unbounded or the maximum number of iterations is reached, we set $\SR\left(\Pos, \EForce[\hat]\right) = \infty$.
We perform experiments with square and octagonal approximations of the friction circle.
Every contact interface is discretized into 20 discrete contact points whose positions are selected using an approximate farthest point sampling strategy where the point, from a subset of 10 randomly selected points, farthest from any previously selected contact point is chosen.

\subsubsection{Approximate}
\label{sec:validation_benchmark_approximate}
The approximate robustness assessment proposed in \cite{nadeau_planning_2024}, in which every object is treated in isolation, is also used as a comparison point.
Contact interfaces are discretized into 20 contact points and contact forces are computed by solving the optimization problem in \cref{sec:solving_forces_problem}.

\subsubsection{Ground Truth}
\label{sec:validation_benchmark_ground_truth}
For simple scenes, it is feasible to manually compute the robustness of the assembly to external forces.
This is done through a geometrical analysis of the assembly and by reasoning about contact forces.
To do so, the surface of the objects onto which external forces are applied is divided into areas. 
For each area, the force necessary to produce slipping is computed, and the axes about which the assembly would topple under the influence of the external force are determined.
The torque necessary to produce toppling about each axis is determined by computing the average moment arm of contact interfaces and considering the forces at the interfaces.
Compared to other approaches, the result thereby obtained is not subject to errors due to contact interface detection or discretization.

\subsubsection{This Work}
\label{sec:validation_benchmark_this_work}
Our proposed robustness assessment method relies on a discretization of contact interfaces into 20 contact points, selected through farthest point sampling, where contact forces are applied. 
This is the same procedure used for the other methods we benchmark.
Areas of contact between objects are detected with a collision detection algorithm. %

\subsection{Evaluation Criteria}
\label{sec:validation_benchmark_evaluation}
We evaluate the performance of the methods in terms of running time and accuracy.
Running time includes the time spent detecting contact areas, computing contact forces, and computing the robustness of the assembly.
Accuracy is evaluated by comparing the robustness obtained with each method to the manually computed ground truth and expressing the difference as a relative error.
Points for which the robustness is infinite are not considered in the accuracy evaluation as they result in an undefined relative error.

\subsection{Results}
\label{sec:validation_benchmark_results}
Ten experiments are performed for each method/assembly combination, for a total of 120 runs.
The running time and accuracy results are averaged over the runs and are shown in \cref{tab:validation_results}.
For the optimization-based method, the results are shown for two different polygonal approximations of the friction cone, with 4 and 8 sides, as each type of approximation yields different results.

\begin{table}[h]
\centering
\begin{tabular}{l|ccc}
\toprule
\textbf{Method / Scene} & \textbf{Cube} & \textbf{Stack} & \textbf{Table} \\
\midrule
\textbf{Approximate} & & & \\
~~Time (s) & \textbf{0.07} & 0.23 & 0.24 \\
~~Error (\%) & \textbf{0} & 30 & 44 \\
\textbf{Simulation-based} & & & \\
~~Time (s) & 0.80 & 10.1 & 41.7 \\
~~Error (\%) & 22 & 26 & 67 \\
\textbf{Optimization-based (4)} & & & \\
~~Time (s) & 10.9 & 151 & 118 \\
~~Error (\%) & 16 & 17 & 23.6 \\
\textbf{Optimization-based (8)} & & & \\
~~Time (s) & 23.6 & 205 & 118 \\
~~Error (\%) & 11 & 10 & 9 \\
\textbf{This Work} & & & \\
~~Time (s) & 0.13 & \textbf{0.20} & \textbf{0.23} \\
~~Error (\%) & \textbf{0.00} & \textbf{1.00} & \textbf{1.20}\\
\bottomrule
\end{tabular}
\caption{Accuracy and running time comparison between several methods used to compute assembly robustness. The benchmark is done on three scenes for which the ground truth robustness is manually computed. Numbers between parentheses indicate the number of sides of the polygonal friction cone approximation used.}
\label{tab:validation_results}
\end{table}

\section{Applications}
\label{sec:applications}
In the following, we demonstrate how our robustness assessment method can be used for stable object placement planning, safe assembly transportation, and disassembly planning.

\subsection{Object Placement Planning}
\label{sec:object_placement_planning}
\begin{figure}
  \centering
  \begin{overpic}[width=1\columnwidth]{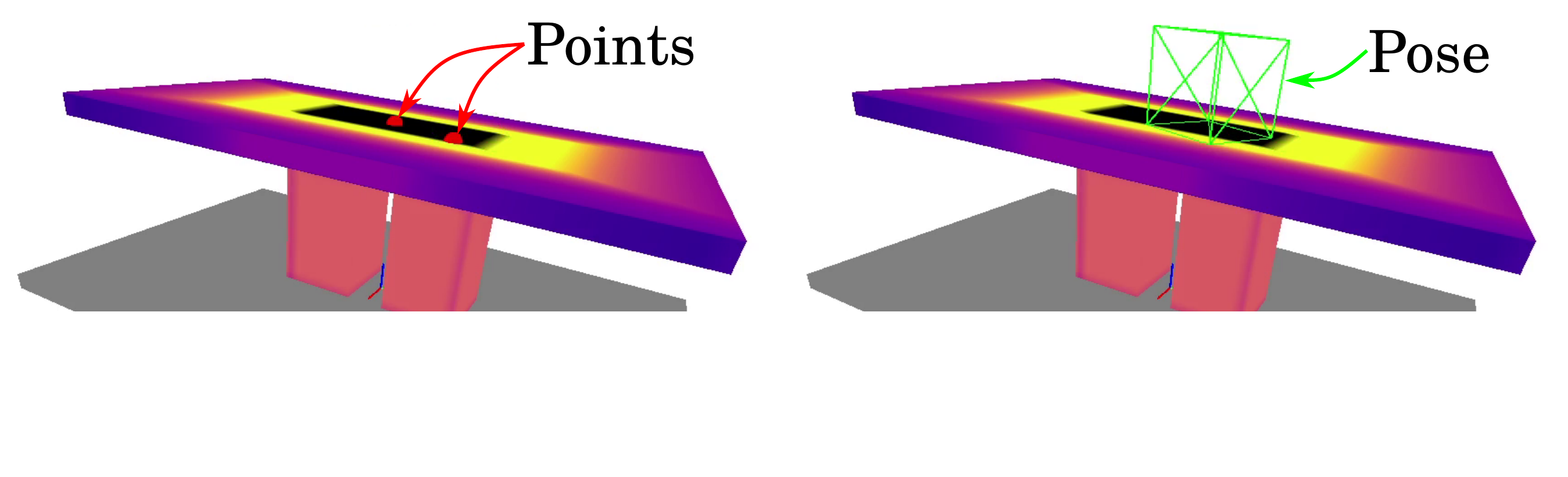}
    \put(5,5){\footnotesize (a) Contact point sampling}
    \put(62, 5){\footnotesize (b) Pose definition}
  \end{overpic}
  \vspace{-8mm}
  \caption{Example of object placement planning from contact point robustness according to \cite{nadeau_planning_2024}. (a) Robustness is used to bias contact point sampling on the assembly. (b) The sampled contact points are used to define a pose for the object to be placed (a small cube).}
  \label{fig:placement_example}
\end{figure}

The problem of stably placing an object in an existing assembly is challenging since feasible poses must (1) make contact with the assembly while avoiding inter-penetrations, and (2) exert forces upon all objects in the assembly while preserving the force equilibrium of the system.
Consequently, a very small region of the solution space is feasible, and searching for stable placements by iterating over a discretized set of poses is inefficient and computationally expensive.
As proposed in \cite{nadeau_planning_2024}, the robustness of points on objects in the assembly can be used to efficiently guide the search for stable poses by sampling robust contact points and determining poses that would solicit the selected contact points.
An example of the overall process is illustrated in \cref{fig:placement_example}.
While the approximate approach in \cite{nadeau_planning_2024} (henceforth referred to as \textit{Approx}) reduced computational complexity by considering each object in isolation, the method proposed in this work can be used to compute the robustness of an assembly more accurately, albeit at a higher computational cost.
Hence, our approach can be seamlessly integrated into the planner proposed in \cite{nadeau_planning_2024} to improve the reliability of the placements found. 

The six benchmarking scenes from \cite{nadeau_planning_2024} (S1--S6) were used to measure the impact our robustness assessment method in a placement planning application.
The scenes range from a simple cube resting on a flat surface to more complex assemblies that require reasoning about orientation and about using multiple objects as support.
Our proposed algorithm was evaluated as a drop-in replacement for \textit{Approx} by reusing the same planning parameters but computing the assembly robustness with the method proposed in this work.
A total of 1,200 experiments were performed across the six scenes and two methods, with the average evaluation criteria of each scene-method pair indicated in \cref{tab:placement_approx_comparison}.
For each experiment, the time taken to find a stable placement was recorded, and the minimal robustness of the assembly following the placement was computed to evaluate the quality of the placement.
\begin{table}
    \centering
    \begin{tabular}[t]{l|c|c|c|c|c|c|c}
        \toprule
        & \textbf{S1} & \textbf{S2} & \textbf{S3} & \textbf{S4} & \textbf{S5} & \textbf{S6} & \textbf{Avg.}\\
        \midrule
      \textbf{Approx} & & & & & & & \\ 
        \hspace{2mm}Time (s) & \textbf{0.387} & 2.28 & 0.551 & \textbf{2.94} & \textbf{2.43} & 1.90 & \textbf{1.75} \\ 
        \hspace{2mm}Rob. (N) & 2.43 & 1.85 & \textbf{0.817} & 0.416 & 1.53 & 2.28 & 1.55 \\ 
      \textbf{Ours} & & & & & & & \\ 
        \hspace{2mm}Time (s) & 0.41 & \textbf{1.95} & \textbf{0.550} & 3.26 & 2.73 & \textbf{1.84} & 1.79 \\ 
        \hspace{2mm}Rob. (N) & \textbf{2.45} & \textbf{2.14} & 0.801 & \textbf{0.478} & \textbf{1.64} & \textbf{2.47} & \textbf{1.66} \\ 
        \midrule
      \textbf{Relative} & & & & & & & \\ 
        \hspace{2mm}Time (\%) & +6 & \textbf{-- 14} & + 0 & + 11 & + 12 & \textbf{-- 3} & + 2 \\  
        \hspace{2mm}Rob. (\%) & \textbf{+ 1} & \textbf{+ 16} & -- 2 & \textbf{+ 15} & \textbf{+ 7} & \textbf{+ 8} & \textbf{+ 7} \\
        \bottomrule
    \end{tabular}
    \caption{Placement planning performance comparison between \cite{nadeau_planning_2024} (Approx) and this work (Ours) on their six benchmarking scenes. The relative improvement of our method is also indicated (bottom).}
    \label{tab:placement_approx_comparison}
\end{table}

\begin{figure}
    \centering
    \subfloat[Approx.]{
        \includegraphics[width=0.4\columnwidth]{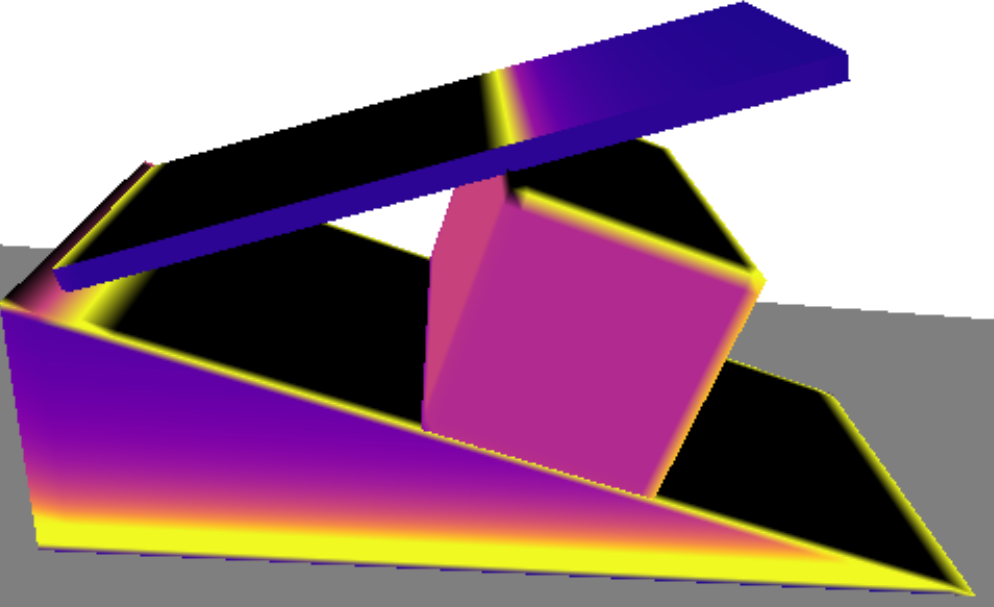}
        \label{fig:slab_cube_slant_approx}
    }
    \subfloat[Ours]{
        \includegraphics[width=0.4\columnwidth]{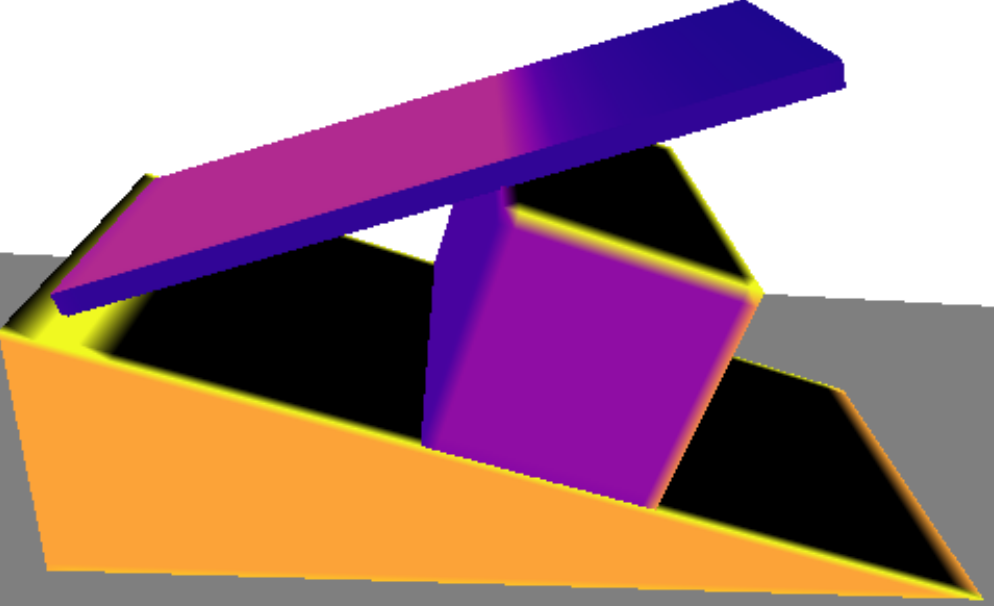}
        \label{fig:slab_cube_slant_exact}
    }\\
    \subfloat[Approx.]{
        \includegraphics[width=0.4\columnwidth]{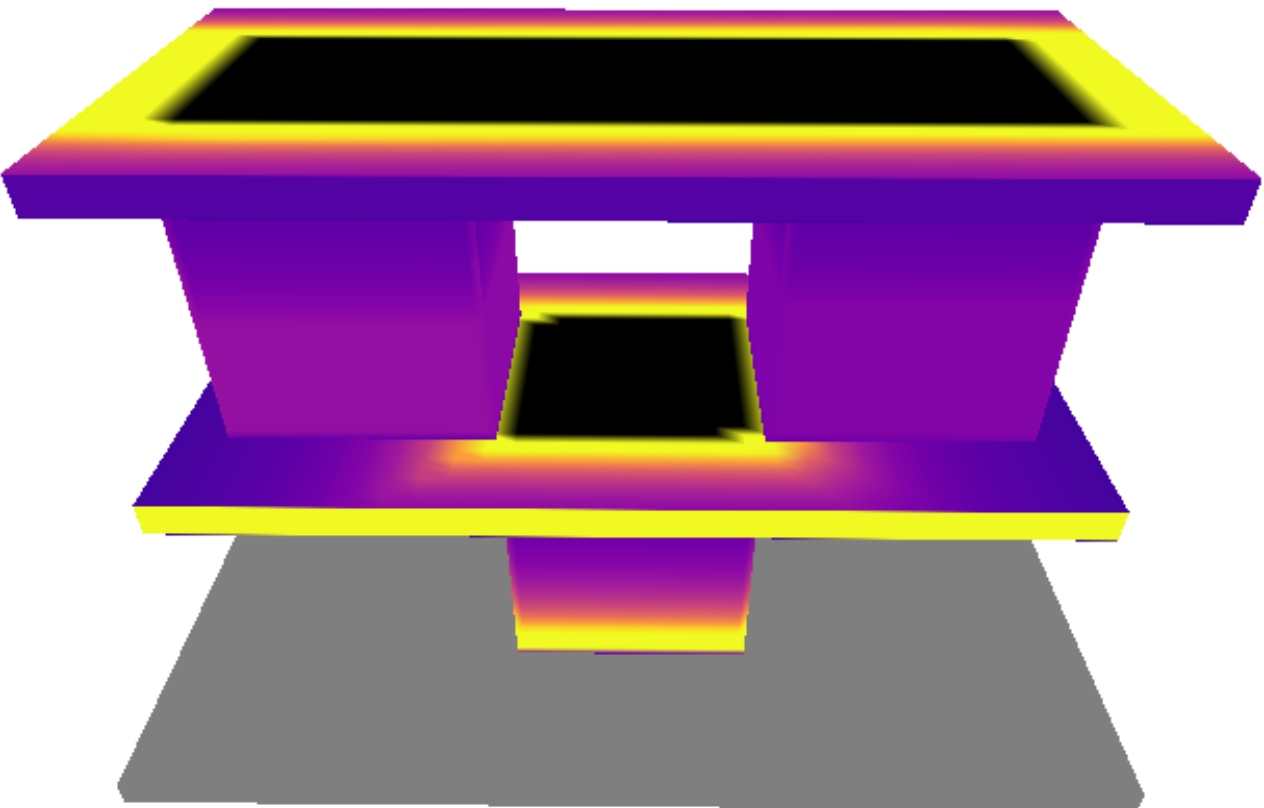}
        \label{fig:double_table_approx}
    }
    \subfloat[Ours]{
        \includegraphics[width=0.4\columnwidth]{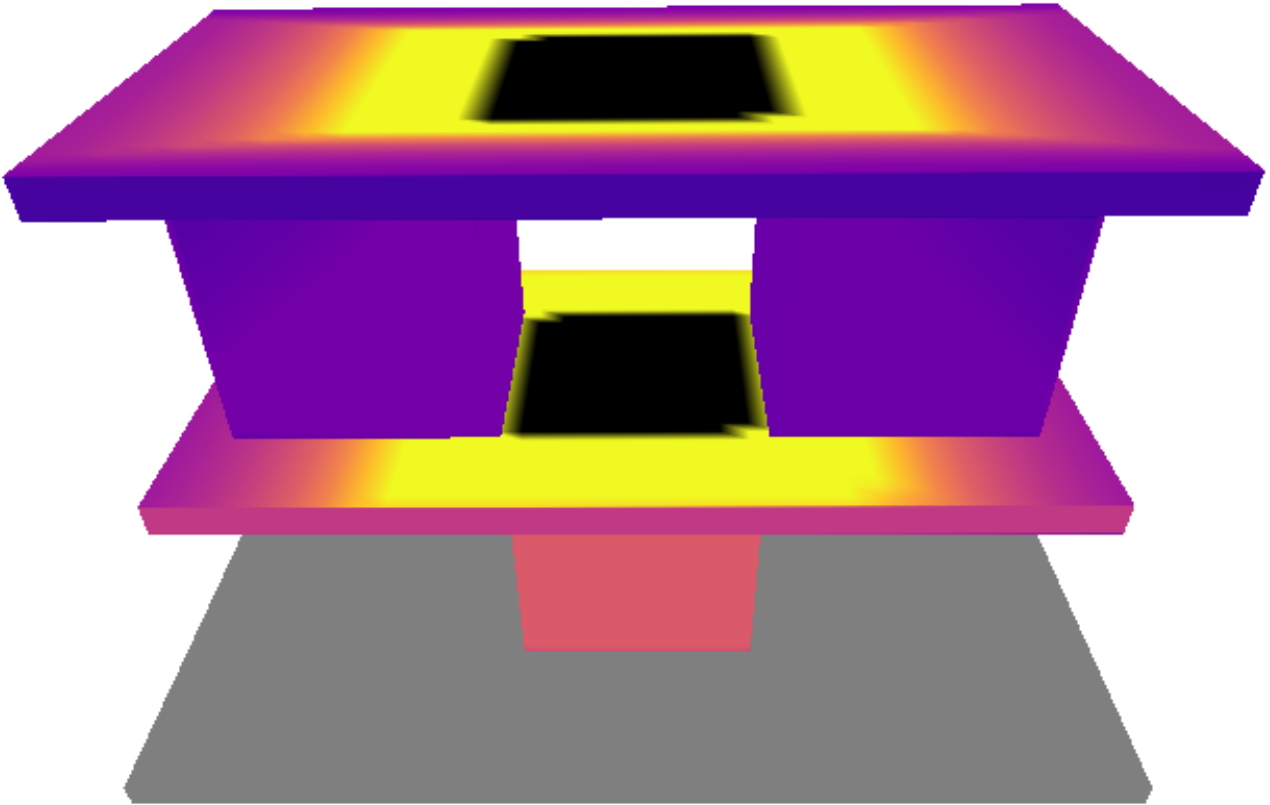}
        \label{fig:double_table_exact}
    }\\
    \subfloat[Approx.]{
        \includegraphics[width=0.4\columnwidth]{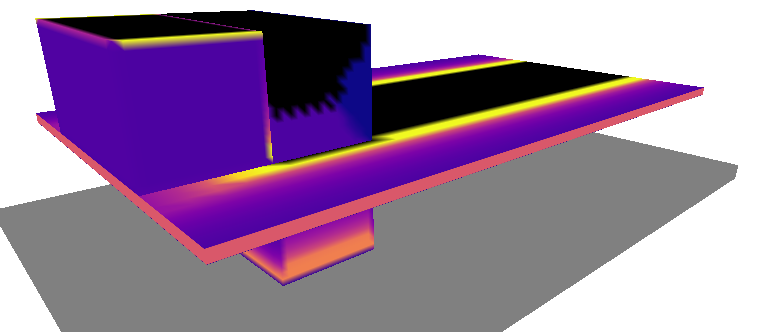}
        \label{fig:cantilever_approx}
    }
    \subfloat[Ours]{
        \includegraphics[width=0.4\columnwidth]{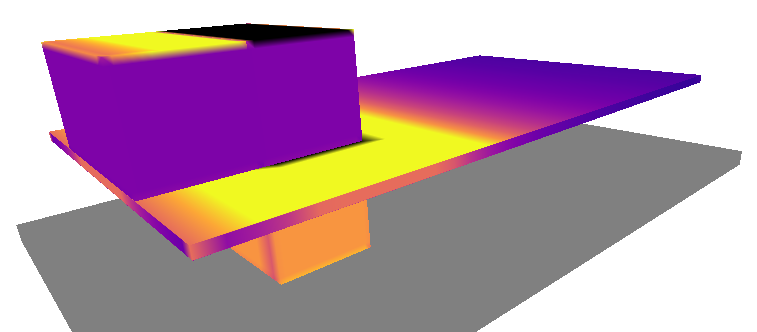}
        \label{fig:cantilever_exact}
    }\\
    \vspace{1mm}
    \begin{overpic}[width=0.9\columnwidth]{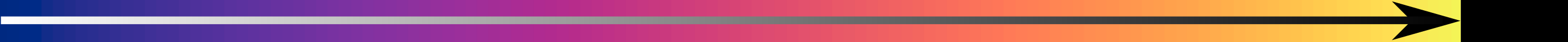}
        \put(-3,0){0}
        \put(94.5,0.5){\color{white}$\infty$}
    \end{overpic}
    \caption{The approximate method (left side) produces robustness maps that can misguide the planner into selecting unstable placements. Our method (right side) is more accurate and can better guide the planner.}
    \label{fig:sr_approx_drawbacks} 
\end{figure}

We also benchmark our proposed method against \textit{Approx} on three additional scenes, shown in \cref{fig:sr_approx_drawbacks}, for which stable placement planning requires reasoning about the complete assembly.
We refer to scenes in the top, middle, and bottom rows of \cref{fig:sr_approx_drawbacks} as \textit{Slant}, \textit{Balance}, and \textit{Cantilever}, respectively.
We perform 100 placement planning experiments on each scene, with the same methods and parameters as in the previous experiments, for a total of 600 runs, and we report the average results in \cref{tab:placement_approx_comparison_newscenes}.
\begin{table}
    \centering
    \begin{tabular}[t]{l|c|c|c|c}
        \toprule
        & \textbf{Slant} & \textbf{Balance} & \textbf{Cantilever} & \textbf{Avg.}\\
        \midrule
      \textbf{Approx.} & & & &\\ 
        \hspace{2mm}Time (s) & 4.5 & 3.0 & 0.7 & 2.7\\ 
        \hspace{2mm}Rob. (N) & 0.3 & 0.5 & 0.5 & 0.4\\ 
      \textbf{Ours} & & & &\\ 
        \hspace{2mm}Time (s) & 3.9 & 1.1 & 0.4 & 1.8\\ 
        \hspace{2mm}Rob. (N) & 0.4 & 0.7 & 0.5 & 0.5\\
        \midrule
      \textbf{Relative} & & & & \\ 
        \hspace{2mm}Time (\%) & \textbf{-- 13} & \textbf{-- 63} & \textbf{-- 43} & \textbf{-- 40}\\ 
        \hspace{2mm}Rob. (\%) & \textbf{+ 33} & \textbf{+ 40} & + 0 & \textbf{+ 24}\\ 
        \bottomrule
    \end{tabular}
    \caption{Results from placement planning experiments on our three more complex scenes from robustness obtained with \cite{nadeau_planning_2024} (\textit{Approx}) and with our method. The relative improvement of our method is also indicated (bottom).}
    \label{tab:placement_approx_comparison_newscenes}
\end{table}

\subsection{Stable Assembly Transportation}
\label{sec:efficient_transportation}
A mobile robot transporting a set of objects, while aiming to minimize the transportation time, should also avoid accelerating too quickly to prevent any object from being destabilized.
Hence, determining the acceleration for which the assembly is on the verge of moving relative to the support---the maximal sustainable acceleration(MSA)---is key to ensuring the safety and efficiency of the transport.
We propose a method that makes use of our robustness criterion to compute the MSA, and we compare our results to simulation-based and optimization-based methods with two simple examples.

According to D'Alembert's principle \cite{goldstein_classical_2002}, an accelerating assembly can be treated as an equivalent static system, with a fictitious force applied at the center of mass in the direction opposite to the acceleration vector.
The MSA of an assembly can be determined from its robustness with \cref{algo:msa}, by iterating over all feasible cuts of the CIG and computing the MSA along desired acceleration vectors for each cut.
The cut with the smallest MSA will be the first to move when the assembly is accelerated in the direction of the corresponding acceleration vector.

\begin{figure}
  \centering
  \begin{overpic}[width=1\columnwidth]{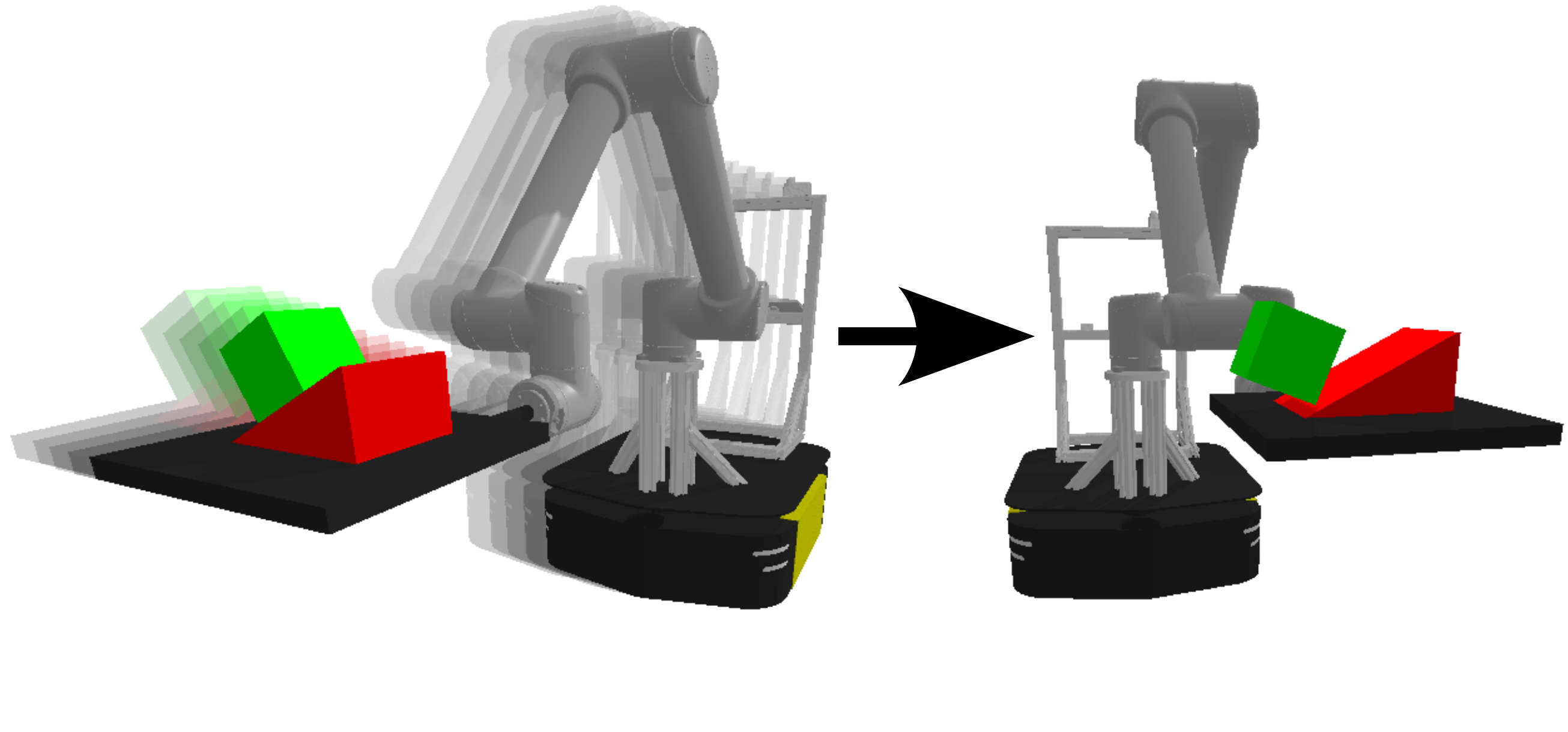}
    \put(10,3){\footnotesize (a) Mobile robot accelerating}
    \put(65,3){\footnotesize (b) MSA Exceeded}
  \end{overpic}
  \caption{An example from our simulations illustrating a mobile robot (a) accelerating below the MSA and (b) exceeding the MSA when travelling in the $+Y$ direction.}
  \label{fig:mobile_robot_accelerating}
\end{figure}

\begin{figure}
    \centering
    \begin{overpic}
        [width=1\columnwidth]{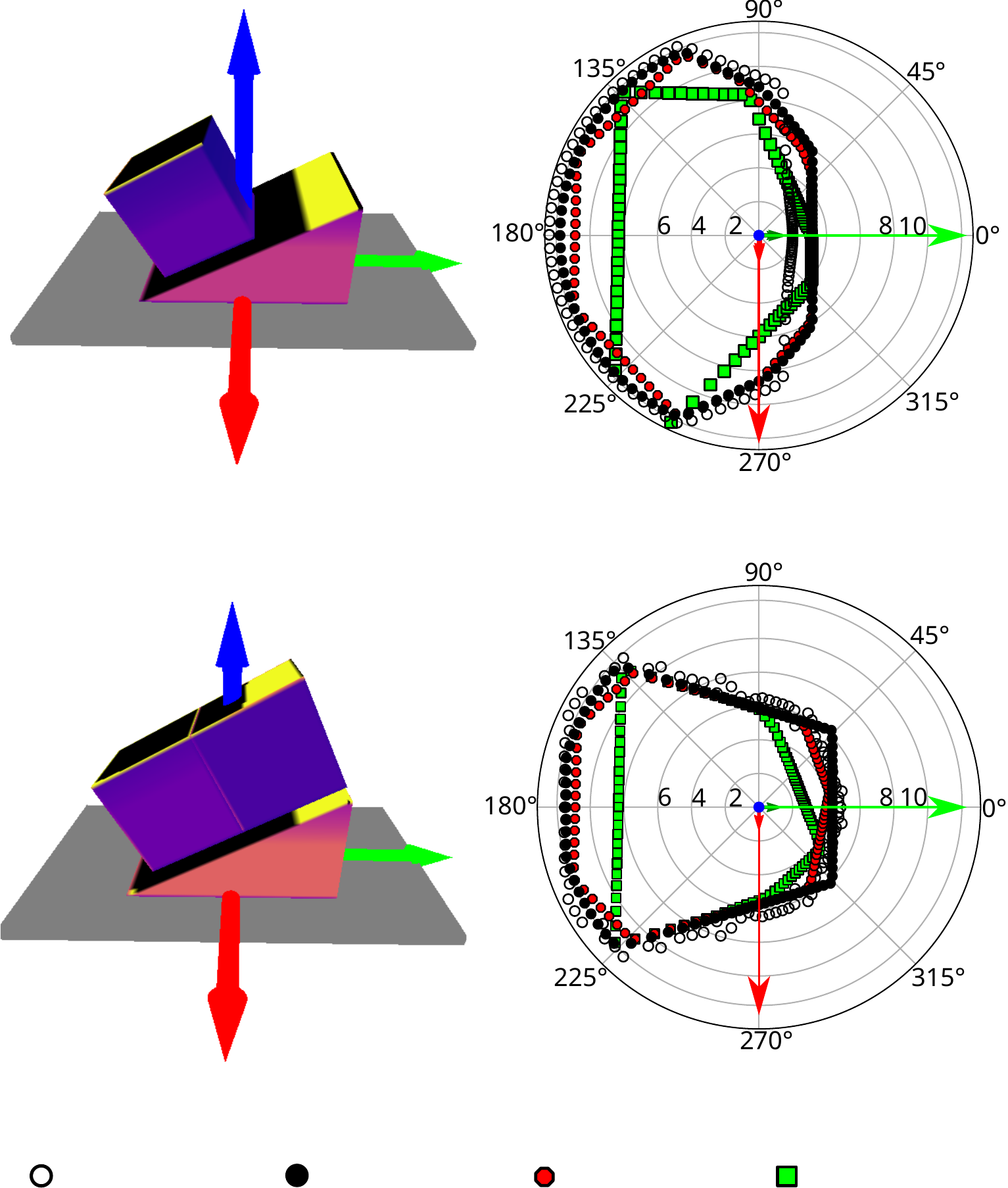}
        \put(40,56.5){\footnotesize (a)}
        \put(40,7){\footnotesize (b)}
        \put(4.9,0){Simulation}
        \put(26.3,0){Robustness}
        \put(47.2,0){Optim. (8)}
        \put(67.4,0){Optim. (4)}
        \put(35,80){+Y}
        \put(35,30){+Y}
        \put(21,61){+X}
        \put(21,11){+X}
    \end{overpic}
    \caption{The maximum sustainable acceleration (in $\rm{m}/\rm{s}^2$) of an assembly in the horizontal plane (right side) when a single cube is on the slant (top), and when a second cube is placed behind the first one (bottom). Filled dots result from our algorithm while empty dots result from dynamics simulations. Red octagons and green squares respectively result from an optimization-based approach with octagonal and square Coulomb circle approximations.}
    \label{fig:msa_example}
\end{figure}

Our algorithm is showcased in \cref{fig:msa_example}, in which the MSA of two assemblies is computed for 100 acceleration vectors in the horizontal plane.
The result from our algorithm is compared to the MSA computed using the simulation environment from \cite{heins_2023_keep}, in which a mobile manipulator carries an assembly as shown in \cref{fig:mobile_robot_accelerating}.
Simulation experiments are performed with the PyBullet physics engine \cite{coumans_2021} at 900 Hz, with a contact error reduction parameter set to 0.2 for \hyperref[fig:msa_example]{\cref*{fig:msa_example}a} and 0.15 for \hyperref[fig:msa_example]{\cref*{fig:msa_example}b}.
The MSA is assumed to be reached when a relative velocity of 0.05 $\rm{m}/\rm{s}$ between objects in the assembly is exceeded.
We also compare to two variations of the optimization-based approach defined in \cref{sec:validation_benchmark_optimization}: one with a square approximation of the friction circle (Optim. (4)) and one with an octagonal approximation (Optim. (8)).
For each acceleration direction, the optimization problem in \cref{eq:opt_based_robustness_start}--\cref{eq:opt_based_robustness_end} is solved with $\UnitVec{e}$ set to the acceleration direction and applied at the centre of mass of each object. 
The time taken to compute the MSA with the simulation-based approach, optimization-based approach, and robustness assessment method for the example in \hyperref[fig:msa_example]{\cref*{fig:msa_example}b} is 38, 3.6 and 0.63 seconds, respectively.

In the example pictured in \hyperref[fig:msa_example]{\cref*{fig:msa_example}a}, the assembly can sustain a large acceleration in the $-Y$ direction (i.e., to the left) due to large coefficients of friction and due to the fact that the inertial force exerted on the cube will tend to push it against the slant and improve the contact conditions.
At some point (i.e., 11.8 $\rm{m}/\rm{s}^2$), however, the inertial force will overcome the frictional forces and the assembly will slide to the right.
In comparison, a much smaller acceleration (i.e., 3.27 $\rm{m}/\rm{s}^2$) can be sustained in the $+Y$ direction before the cube topples to the left under the influence of the inertial force.
Adding a second cube behind the first one, as shown in \hyperref[fig:msa_example]{\cref*{fig:msa_example}b}, prevents the first cube from toppling to the left and gives rise to a 30\% MSA increase in the $+Y$ direction.
This comes at the cost of a MSA decrease in the $\pm X$ directions, as the second cube elevates the centre of mass of the assembly.

In practice, the MSA can be used to inform kinodynamic motion planners \cite{donald_1993_kinodynamic} on the maximal accelerations that can be safely reached by a mobile robot transporting an assembly \cite{kunz_2014_probabilistically, lavalle_2001_randomized, webb_2013_kinodynamic, lau_2009_kinodynamic}.

\begin{algorithm}
    \caption{Maximal Sustainable Accelerations}
    \label{algo:msa}
    \DontPrintSemicolon
    \SetKwInOut{Input}{input}\SetKwInOut{Output}{output}
    \Input{Set of 3D acceleration directions $\mathcal{D}$}
    \Output{Set of maximal sustainable accelerations $\mathcal{X}$}
    1. Initialize all $x_i \in \mathcal{X}$ to $\infty$\;
    2. Determine the feasible CIG cut set $\mathcal{C}$\;
    \ForEach{$c \in \mathcal{C}$}{
        3. Compute the mass $\Mass$ and centre of mass $\CenterOfMass$ of $c$\;
        \ForEach{$\Vector{d}_i \in \mathcal{D}$}{
            4. Compute the MSA in $\Vector{d}_i$ with $x_i \gets \min\left\{x_i, \SR(\CenterOfMass, -\Vector{d}_i)/\Mass\right\}$\;
        }
    }
\end{algorithm}

\subsection{Disassembly Planning}
\label{sec:safe_disassembly_planning}
\begin{figure}
  \centering
  \includegraphics[width=0.8\columnwidth]{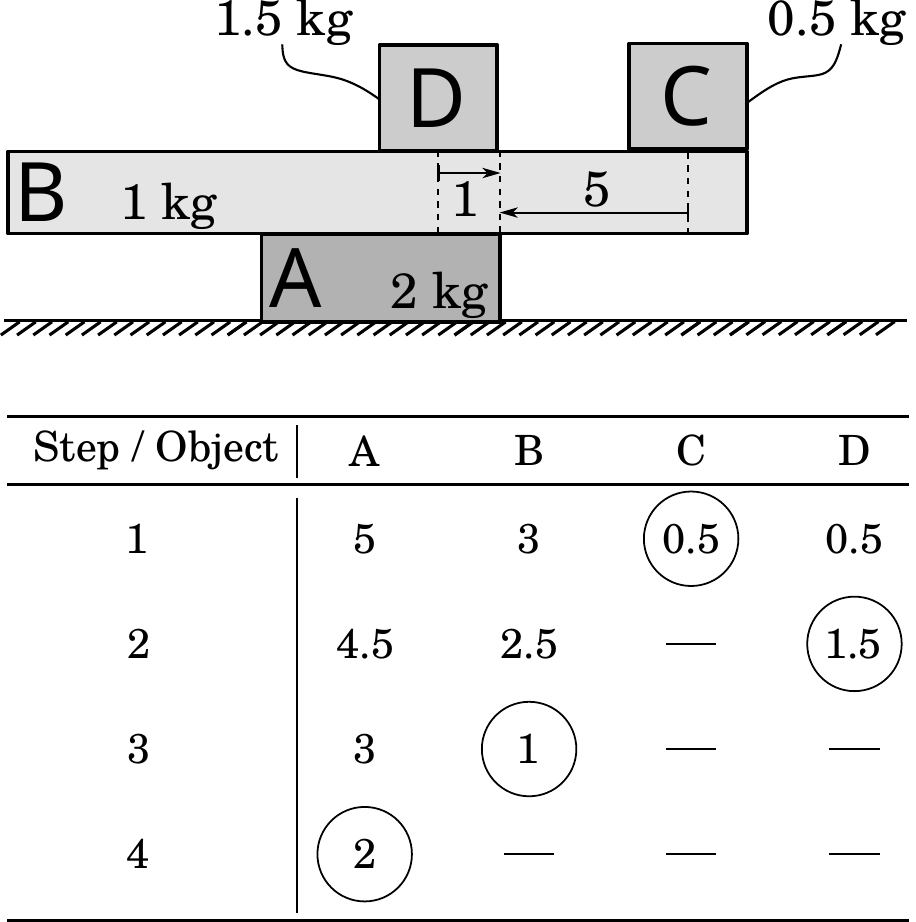}
  \caption{Disassembly planning of a structure of four objects (top). The order in which objects can be safely removed is determined by iteratively computing the robustness of the assembly to forces exerted at the centre of mass of each object and directed opposite to their acceleration (table values). An object can be safely removed when this value is equal to its weight. In this example, the disassembly order is: C, D, B, A.}
  \label{fig:disassembly_planning_example}
\end{figure}

Determining the order in which objects in contact can be removed without causing instabilities in the assembly can enable robots to automate disassembly operations that are requisite when a subset of the objects in the assembly have to be moved.
The tasks of disassembly sequencing and backward assembly planning (i.e., assembly by disassembly), which recursively decompose a given assembly into simpler subassemblies, can be solved by considering cuts in the kinematic structure \cite{lee_1994_forceflownetwork, lee_1999_disassembly}. 
Since our proposed approach is based on a graph representation of the assembly (i.e., the CIG), it can directly be used for disassembly planning.
To this end, the robustness of the assembly to forces exerted at the centre of mass of each object and directed opposite to their acceleration (i.e., gravity) is computed.
An object can be safely removed when the force required to move it is strictly opposed by its inertia (e.g., its weight).
A lower or higher robustness at the centre of mass would indicate that the assembly depends on the object to remain stable.
An example of disassembly planning is shown in \cref{fig:disassembly_planning_example}, where the order in which objects can be safely removed is determined by iteratively computing the robustness of the assembly to forces exerted at the centre of mass of each object and directed opposite to gravity (i.e., upward).

\section{Discussion}
\label{sec:discussion}

\subsection{Robustness Assessment: Sources of Errors}
\label{sec:discussion_on_robustness_assessment}
In this work, our proposed method is compared to three other methods used for computing the robustness of assemblies: an optimization-based approach, a simulation-based approach, and a heuristic.
All methods rely on discretizing contact interfaces, thereby violating the assumption of uniform pressure distribution and leading to erroneous estimates of the centre of friction estimate.
Nonetheless, contact interface discretization is reasonable with rigid objects whose coefficient of friction has been shown to vary significantly across contact areas \cite{yu_more_2016}, thereby introducing uncertainty in the location of the centre of friction. 

According to our validation results in \cref{tab:validation_results}, the simulation-based approach yields a large average relative robustness error of 38\% on simple scenes.
This can be explained by the fact that dynamics simulations do not assume static equilibrium, consider few contact points at every time step, and make various approximations to speed up the process.

In contrast, optimization-based approaches can yield more accurate results but are limited by the Coulomb friction circle approximation.
On average, the distance between any point on a circle and its closest point on an inscribed square is about 20\% of the circle radius.
For an inscribed octagon, the average distance is about 10\% of the circle radius.
These values correspond to the average relative robustness errors obtained with the square and octagonal approximations in \cref{tab:validation_results}, suggesting that optimization-based approaches can yield accurate robustness estimates when using a large number of sides to approximate the Coulomb circle.

Our proposed method solves the quadratic equation in \cref{eq:single_contact_slipping_condition} when computing robustness instead of relying on linear approximations of the Coulomb circle.
However, like the other approaches, the accuracy of our method is limited by the discretization of contact interfaces, which is likely responsible for the small relative robustness errors shown in \cref{tab:validation_results}.

Finally, like our method, the approximation proposed in \cite{nadeau_planning_2024} does not rely on linear approximations of the Coulomb circle and yields accurate results for isolated objects. 
However, for assemblies whose CIG has a depth greater than one, the approximate method will be inaccurate, as suggested by our validation results.

\subsection{Performance in Object Handling Applications}
\label{sec:discussion_on_performance_in_object_handling_applications}
Assessing the robustness of assemblies to external forces can be used in various object handling operations, such as object placement, safe transportation, and disassembly planning.
In \cref{sec:applications}, we provided examples of how our method can be used for these purposes.
This section discusses the performance of our method in these applications.

Comparing the performance of the object placement planner proposed in \cite{nadeau_planning_2024} (Approx) to a variation of the same planner that uses our robustness assessment method, we observe that our method is competitive in terms of planning time (+2\%) and produces slightly more stable placements (+7\%).
On more complex scenes, however, results in \cref{tab:placement_approx_comparison_newscenes} suggest that using our method can significantly improve the planning time while yielding more robust placements.
On three examples of such scenes, our method is 40\% faster in finding a stable placement and yields assemblies that are 24\% more robust on average.
Although the computational complexity of our robustness assessment method exceeds that of Approx, these results can be explained by the fact that the robustness assessment only has to be performed once per placement while a large number of placement planning iterations can be required to find a stable placement.

For efficiently transporting assemblies, our results in \cref{fig:msa_example} clearly illustrate the importance of carefully selecting the orientation of the assembly when transporting it.
While simulation-based or optimization-based approaches can be used to compute the maximal sustainable acceleration, our method does not require parameter tuning, is faster, and more accurate.

Finally, we also showed how an accurate robustness assessment can be used to determine the order in which objects can be safely removed from an assembly, enabling planning disassembly sequences.

\subsection{Practical Considerations: Approximations and Safety}
\label{sec:discussion_on_practical_considerations}
The worst-case running time of \cref{algo:toppling_robustness} is proportional to the number of feasible cuts in the CIG and exponential in the number of objects in the assembly.
A practical approach to limit the running time of our method consists in considering that heavy objects are in fact immobile.
This can greatly reduce the number of feasible cuts in the CIG and lower the computational complexity.
In practice, the CIG can be easily analyzed to determine whether such an approximation is needed or applicable.

When assembling or transporting objects, it might be beneficial to reduce the odds of instability occurring due to object model inaccuracies. 
A simple strategy to achieve this consists in considering that the mass of objects is smaller than their estimated value.
Since robustness linearly increases with object mass, underestimating the mass of objects will lead to a conservative estimate of the assembly robustness.

\subsection{Limitations and Future Work}
\label{sec:discussion_on_limitations_and_future_work}
Our proposed method for computing the robustness of assemblies to external forces assumes that object pose, friction coefficients, and object shapes are known with certainty. 
While object shapes can be accurately provided by CAD models, practical applications will require estimating the pose of objects in the assembly.
Furthermore, friction coefficients have been shown to vary significantly across contact areas \cite{yu_more_2016}.
Future work could consider propagating uncertainty from object poses and friction coefficients into the robustness assessment, thereby enabling robust and reliable object handling operations.
Moreover, while our method assumes that a single external force is exerted on the assembly, bimanual manipulation tasks may require considering the influence of multiple external forces acting on the assembly.
Considering the influence of multiple simultaneous external forces on the assembly is an interesting avenue for future work.

\section{Conclusion}
\label{sec:conclusion}
In this work, we proposed a method to compute the robustness of an assembly to external forces, facilitating autonomous decision-making in tasks involving rigid objects in contact.
Our method does not rely on heuristics or approximations, making it dependable in a broad range of scenarios involving multi-object assemblies.
Through theoretical validation and experimental assessment, we demonstrated that our method is more accurate and much more efficient than existing approaches for computing the robustness of assemblies.
We showed that our method can be used to plan stable object placements in complex assemblies, compute the maximal sustainable acceleration of mobile robots transporting assemblies, and plan disassembly sequences.
We expect our algorithms to be valuable in various applications involving forceful interactions with the environment, particularly in the context of autonomous object handling.

\bibliographystyle{ieeetr}
\bibliography{references}

\end{document}

%% file: notation-shortcuts.tex
\usepackage{xparse}
\usepackage{tikz}

\newcommand{\lrss}[5]{%
	\setbox1=\hbox{\ensuremath{^{#1}}}%
	\setbox2=\hbox{\ensuremath{_{#2}}}%
	\setbox5=\hbox{\ensuremath{#5}}%
	\setbox6=\hbox{\ensuremath{^{#1#3}}}%
	\setbox7=\hbox{\ensuremath{_{#2#4}}}%
	\setbox8=\hbox{\ensuremath{^{#3}}}%
	\setbox9=\hbox{\ensuremath{_{#4}}}%
	\hspace{\ifnum\wd1>\wd2\wd1\else\wd2\fi}%
	\ensuremath{\copy5%
		^{\hspace{-\wd1}\hspace{\wd1}\hspace{\wd8}%
			\hspace{-\wd6}\hspace{-\wd5}#1\hspace{\wd5}#3}%
		_{\hspace{-\wd2}\hspace{\wd2}\hspace{\wd9}%
			\hspace{-\wd7}\hspace{-\wd5}#2\hspace{\wd5}#4}%
}}

\NewDocumentCommand\bbm{}{ \begin{bmatrix} }
\NewDocumentCommand\ebm{}{ \end{bmatrix} }
\NewDocumentCommand\Vector{m}{ \boldsymbol{\mathbf{#1}} }
\NewDocumentCommand\Matrix{m}{ \boldsymbol{\mathbf{#1}} }
\NewDocumentCommand\UnitVec{m}{ \Vector{\hat{#1}} }

\NewDocumentCommand\Norm{m}{\left\Vert#1\right\Vert }

\NewDocumentCommand\Transpose{}{^\mathsf{T}}

\NewDocumentCommand\Real{}{ \mathbb{R} }
\NewDocumentCommand\NonnegativeReal{}{ \Real_+ }

\NewDocumentCommand\IdentityMatrix{m}{ \Matrix{\IdentitySymbol}_{#1} }

\NewDocumentCommand\CoordinateFrame{m}{ \underrightarrow{\Matrix{\mathcal{F}}}_{#1} }

\NewDocumentCommand\IdentitySymbol{}{ 1 }
\NewDocumentCommand\Skew{m}{\left[#1\right]_\times}

\NewDocumentCommand\DotP{}{ \cdot }
\NewDocumentCommand\CrossP{}{ \times }

\NewDocumentCommand\Mass{}{ m }

\NewDocumentCommand\CenterOfMass{}{ \Vector{c} }

\RenewDocumentCommand\CoordinateFrame{m}{ \Matrix{\mathcal{F}}_{#1} }
\NewDocumentCommand\CFrame{m}{ \CoordinateFrame{#1} }
\NewDocumentCommand\CSys{m}{ \{#1\} }